\documentclass{article} 
\usepackage[preprint]{colm2026_conference}

\usepackage{microtype}
\usepackage{url}
\usepackage{booktabs}
\usepackage{multirow}
\usepackage{makecell} 
\usepackage{graphicx}

\usepackage{xcolor}

\usepackage{lineno}
\usepackage{adjustbox}
\usepackage{booktabs}
\usepackage{longtable}
\usepackage{xcolor}
\usepackage{colortbl}
\usepackage{makecell}
\usepackage{adjustbox}
\usepackage{listings}
\usepackage{hyperref}

\usepackage{fontawesome5}

\definecolor{darkblue}{rgb}{0, 0, 0.5}
\definecolor{darkgreen2}{HTML}{0A6E2E}

\usepackage{microtype}

\usepackage{amsmath}
\usepackage{amssymb}
\usepackage{amsfonts}
\usepackage{amsthm}
\usepackage{mathtools}
\usepackage{bm}        
\usepackage{bbm}       
\usepackage{nicefrac}  

\usepackage{graphicx}
\usepackage{subcaption}

\usepackage{booktabs}
\usepackage{array}
\usepackage[table]{xcolor}

\usepackage[inline]{enumitem}
\usepackage{url}

\usepackage{algorithm}
\usepackage{algpseudocode}

\usepackage[most]{tcolorbox}

\usepackage{natbib}

\usepackage[capitalize,noabbrev]{cleveref}

\usepackage[textsize=tiny]{todonotes}

\usepackage[T1]{fontenc}
\usepackage[utf8]{inputenc}
\usepackage[dvipsnames]{xcolor}
\usepackage{booktabs}
\usepackage{array}
\usepackage{graphicx}
\usepackage{amsmath}
\usepackage{caption}
\usepackage{makecell}
\definecolor{darkgreen}{RGB}{80,200,80}
\definecolor{newblue}{HTML}{0064E0}
\definecolor{nblue}{HTML}{D7E5FF}
\definecolor{ngreen}{HTML}{E7FCF2}
\definecolor{nred}{HTML}{FFE8ED}

\hypersetup{
  colorlinks,
  linkcolor=newblue,
  citecolor=newblue,
  urlcolor=newblue
}

\crefname{section}{Sec.}{Secs.}
\crefname{subsection}{Subsec.}{Subsecs.}
\crefname{figure}{Fig.}{Figs.}
\crefname{table}{Tab.}{Tabs.}
\crefname{equation}{Eq.}{Eqs.}
\crefname{appendix}{App.}{Apps.}

\newtcolorbox[auto counter]{theorembox}[2][]{%
  colframe=nblue, colback=nblue, breakable,
  before upper={{\textbf{Theorem \thetcbcounter.} #2}}, #1
}
\makeatletter
\def\tcb@cnt@theoremboxautorefname{Thm.}
\makeatother
\crefname{tcb@cnt@theorembox}{Thm.}{Thms.}

\newtcolorbox{qoutebox}[1][]{%
  colframe=ngreen, colback=ngreen,
  #1
}

\newtcolorbox[auto counter]{definitionbox}[2][]{%
  colframe=nred, colback=nred, breakable,
  before upper={{\textbf{Definition \thetcbcounter.} #2}}, #1
}
\makeatletter
\def\tcb@cnt@definitionboxautorefname{Def.}
\makeatother
\crefname{tcb@cnt@definitionbox}{Def.}{Defs.}

\newtcolorbox[auto counter]{algbox}[2][]{%
  colframe=nred, colback=nred, left=1pt, right=1pt, top=0.5pt, breakable,
  before upper={{\textbf{Algorithm \thetcbcounter.} #2}}, #1
}
\makeatletter
\def\tcb@cnt@algboxautorefname{Alg.}
\makeatother
\crefname{tcb@cnt@algbox}{Alg.}{Algs.}

\newtcolorbox[auto counter]{mainbox}[2][]{%
  colframe=ngreen,
  colback=ngreen,
  breakable,
  before upper={{\textbf{Main Contributions.} #2}},
  #1
}

\usepackage{amsmath,amsfonts,bm}

\def\eqref#1{equation~\ref{#1}}

\def\1{\bm{1}}

\DeclareMathAlphabet{\mathsfit}{\encodingdefault}{\sfdefault}{m}{sl}
\SetMathAlphabet{\mathsfit}{bold}{\encodingdefault}{\sfdefault}{bx}{n}

\newcommand{\method}{\textsc{SCaTR}}

\title{\method: Simple Calibrated Test-Time Ranking}

\author{Divya Shyamal\thanks{Equal contribution.}\\
\href{mailto:dshyamal@mit.edu}{\texttt{dshyamal@mit.edu}}
\vspace{0.09em}\\
\includegraphics[height=1em]{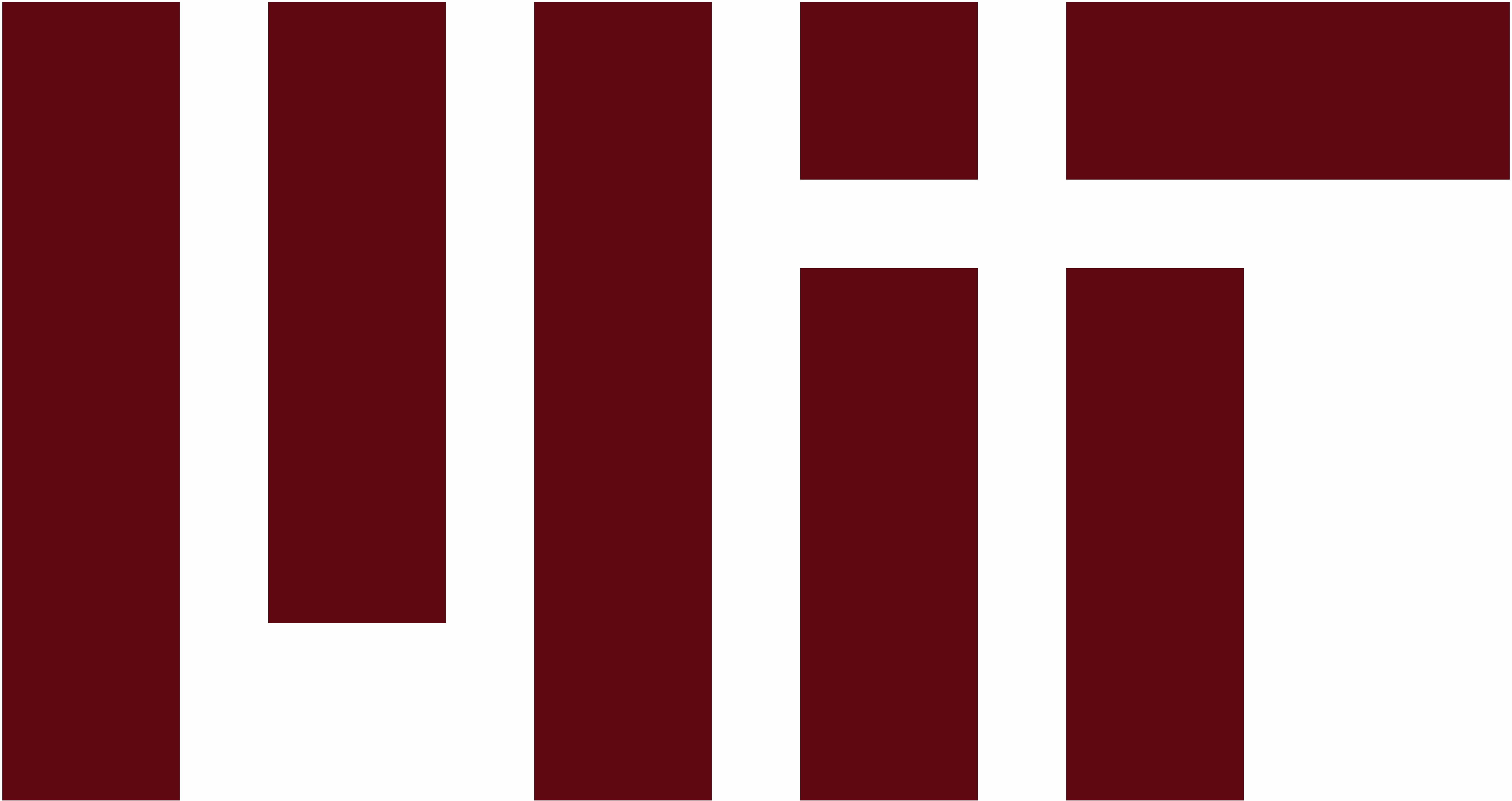}
\And
Marta Knežević\footnotemark[1] \\
\href{mailto:marta.knezevic@epfl.ch}{\texttt{marta.knezevic@epfl.ch}}
\vspace{0.09em}\\
\includegraphics[height=1em]{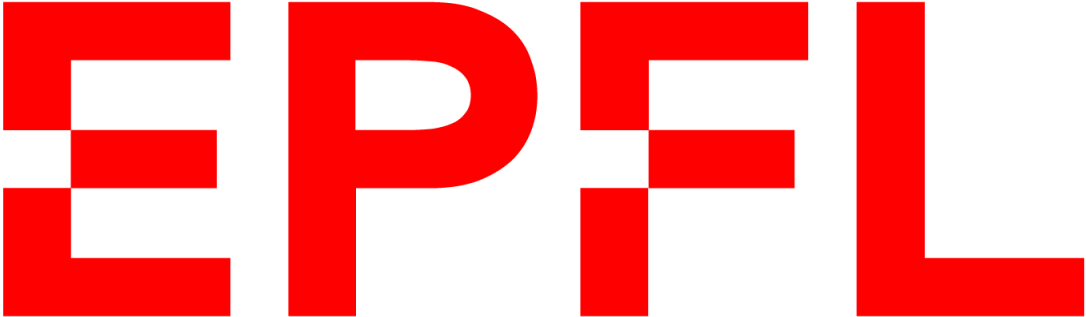}
\And
Lan Tran\footnotemark[1] \\
\href{mailto:tranhuonglantk@gmail.com}{\texttt{lan.tranhuong@epfl.ch}}
\vspace{0.09em}\\
\includegraphics[height=1em]{logos/epfl_logo_cropped_new.png}
\AND
Chanakya Ekbote\thanks{Equal advising.} \\
\href{mailto:cekbote@mit.edu}{\texttt{cekbote@mit.edu}}
\vspace{0.09em}\\
\includegraphics[height=1em]{logos/mit_logo_std_cmyk_mit-red_cropped.jpg}
\And
Vijay Lingam\footnotemark[2]   \textsuperscript{\hspace{0.10em}} \\
\href{mailto:vijaylingam@utexas.edu}{\texttt{vijaylingam@utexas.edu}}
\vspace{0.09em}\\
\includegraphics[height=1.1em]{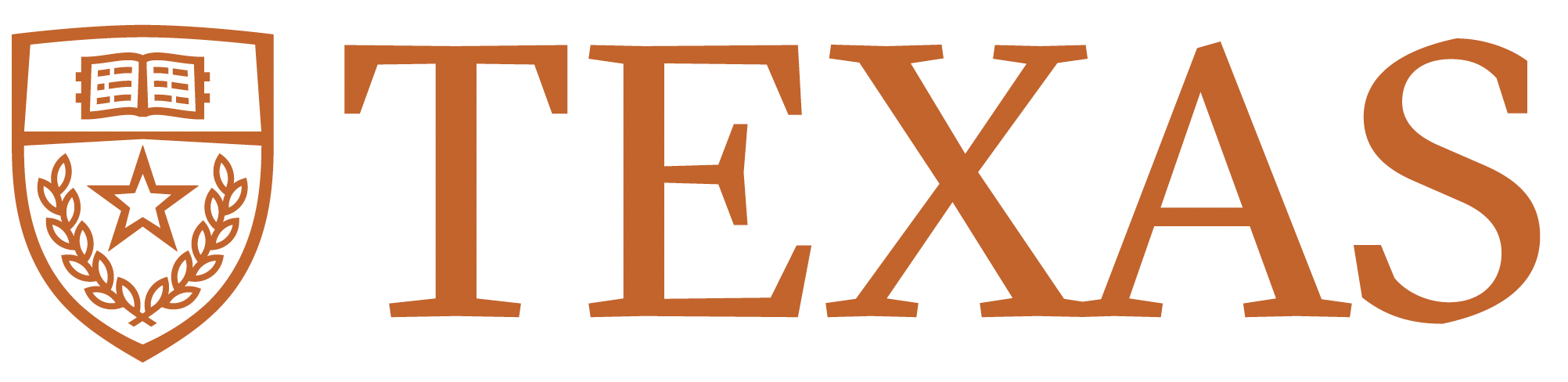}
\And
Paul Pu Liang\footnotemark[2] \\
\href{mailto:ppliang@mit.edu}{\texttt{ppliang@mit.edu}}
\vspace{0.09em}\\
\includegraphics[height=1em]{logos/mit_logo_std_cmyk_mit-red_cropped.jpg}
}

\begin{document}

\ifcolmsubmission
\linenumbers
\fi

\maketitle

\begin{abstract}
Test-time scaling (TTS) improves large language models (LLMs) by allocating additional compute at inference time. In practice, TTS is often achieved through parallel scaling: generating multiple candidate responses and selecting the best via a Best-of-$N$ (BoN) strategy. Its effectiveness therefore hinges on the scoring function. Learned scorers such as process reward models (PRMs) can be strong, but they are expensive to train and run. Lightweight confidence heuristics based on token log-probabilities are much cheaper, yet we find that they often perform substantially worse. To improve on lightweight confidence heuristics without incurring the full cost of stronger learned scorers, we introduce \method, a simple and efficient BoN ranking method that learns a lightweight scorer from a small calibration set using hidden representations from the base model. Across coding and mathematical reasoning benchmarks, \method~improves over prior confidence-based baselines by up to $\mathbf{9\%}$. Relative to LoRA fine-tuning on the same calibration data, it achieves comparable accuracy with up to $\mathbf{8000\times}$ fewer trainable parameters and much lower compute, reducing training and inference latency by up to $\mathbf{150\times}$ and $\mathbf{1000\times}$, respectively. \method~is also competitive with strong PRM baselines, and in several settings improves accuracy by up to $\mathbf{7.8\%}$ on math and $\mathbf{4.2\%}$ on coding while enabling up to $\mathbf{1000\times}$ faster inference. Overall, \method~ offers a strong accuracy-efficiency trade-off for scalable test-time selection. Code is available at \href{https://github.com/martaknezevic/SCaTR}{\faGithub\ this link }.\looseness=-1

\end{abstract}

\section{Introduction}

\begin{qoutebox}
\textit{``Look beneath the surface; let not the several quality of a thing nor its worth escape thee.''} 
\makebox[\linewidth][r]{---\textit{Marcus Aurelius}}
\end{qoutebox}

\begin{figure}[t]
    \centering
    \vspace{2mm}
    \begin{minipage}{0.54\linewidth}
        \centering
        \includegraphics[width=\linewidth]{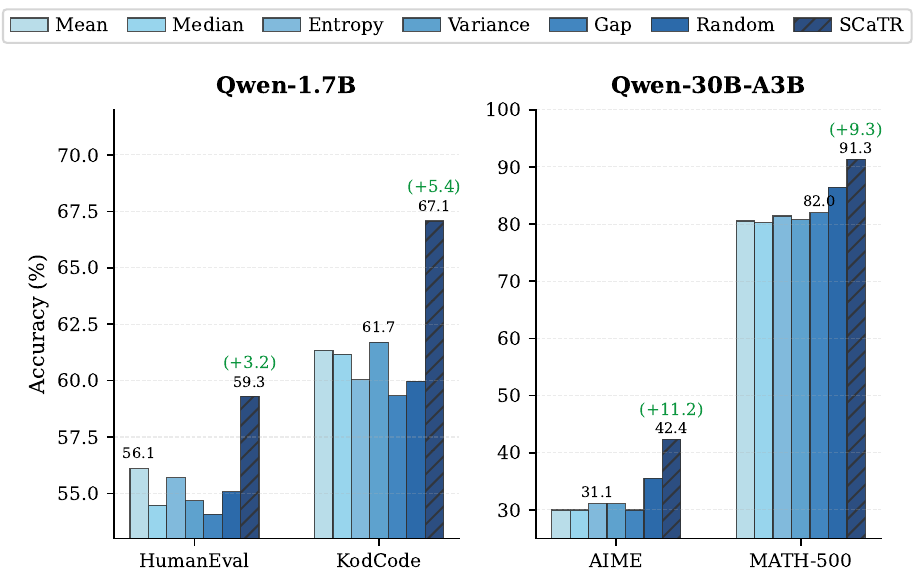}
    \end{minipage}
    \hfill
    \begin{minipage}{0.45\linewidth}
        \centering
        \includegraphics[width=\linewidth]{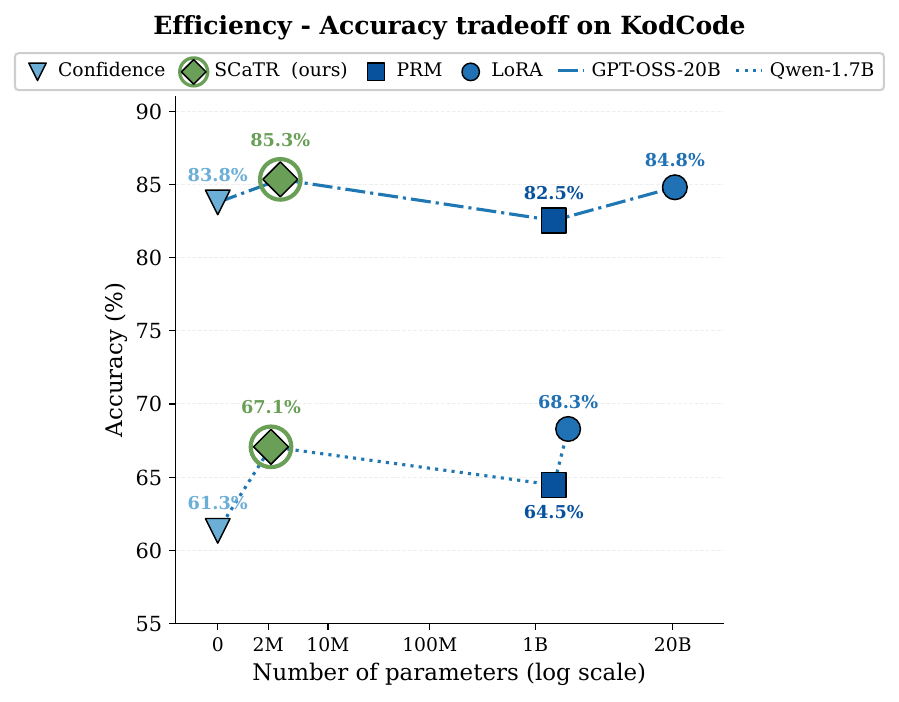}
    \end{minipage}
    \caption{\textbf{Left:} Performance of tail-aggregated token-level uncertainty metrics compared to random best-of-$N$ response selection and \method~. Metrics include standard mean confidence ($C_i$)~\citep{fu2025deepthinkconfidence}, median and variance of top-$k$ log-probabilities, probability gap between the top two tokens, and Shannon entropy~\citep{zhu2026edisdiagnosingllmreasoning, kang2025scalablebestofnselectionlarge}, computed from normalized top-$k$ distributions and aggregated over the sequence tail (see ~\autoref{app:confidence_definitions_and_metrics}). \method~consistently outperforms all confidence-based metrics, which perform close to random selection, highlighting the limitations of fixed uncertainty measures. \textbf{Right:} Selection accuracy across varying model scales (GPT-OSS-20B and Qwen3-1.7B) on the KodCode dataset. \method{} matches or exceeds substantially larger models, demonstrating that lightweight calibration-based scoring can close the gap to higher-capacity systems, making it a compute-efficient approach for TTS. \looseness=-1}
    \label{fig:metric_selection}
\end{figure}

Test-time scaling (TTS) improves language model performance at inference by allocating additional compute, for example through iterative refinement or by sampling multiple candidate responses \citep{snell2024scalingllmtesttimecompute, muennighoff2025s1simpletesttimescaling, ttssurvey}. Prior work has shown that this extra compute can substantially improve performance \citep{fu2025deepthinkconfidence, wang2023selfconsistencyimproveschainthought, chen2023universalselfconsistencylargelanguage}. TTS methods can be broadly divided into sequential and parallel scaling. Sequential scaling \citep{ madaan2023self, yao2023tree, chen2023teaching}  refines a response over multiple dependent generation steps, such as revising intermediate reasoning or correcting earlier errors. While effective, this dependency structure limits throughput because refinement cannot be parallelized. Parallel scaling~\citep{fu2025deepthinkconfidence, kang2025scalablebestofnselectionlarge}, by contrast, generates multiple candidates independently and selects the best one, making it a more attractive approach for scalable inference. This setting reduces TTS to a selection problem: given $N$ sampled candidates, how can we identify the best response? Best-of-$N$ (BoN) has emerged as a standard framework, where candidate responses are scored and the top-ranked output is returned. Existing scoring methods range from aggregation-based approaches such as self-consistency \citep{chen2023universalselfconsistencylargelanguage} to trained reward models \citep{cobbe2021trainingverifierssolvemath, kuang2025optimalaggregationllmprm, zou2025reasonfluxprmtrajectoryawareprmslong, khalifa2025processrewardmodelsthink} and model-based confidence or verification objectives \citep{kadavath2022languagemodelsmostlyknow, chuang2025learningroutellmsconfidence}. Although effective, many of these approaches introduce additional training or deployment cost; Reward models for BoN demand curating training datasets and extensive compute.~\autoref{fig:metric_selection} (left) illustrates the gap between existing confidence based BoN methods and oracle selection.

To avoid this overhead, recent work has explored lightweight confidence metrics derived from final-layer token probabilities \citep{kang2025scalablebestofnselectionlarge, fu2025deepthinkconfidence, chen2023universalselfconsistencylargelanguage}. These methods use token-level softmax confidence or log probabilities as proxies for correctness, offering an efficient alternative to reward modeling. However, such signals can be poorly calibrated: models may be overconfident in incorrect responses and underconfident in correct ones, limiting their reliability for selection \citep{epstein2025llms}. This challenge is especially acute in free-form tasks such as coding or open-ended generation, where candidates often lack a shared answer form and thus do not admit simple aggregation strategies like majority voting (see~\autoref{fig:metric_selection}). These limitations suggest that more informative selection signals may lie beyond output probabilities. Recent work supports this view, showing that hidden states capture rich contextual structure beyond logits alone and can support downstream tasks such as hallucination detection \citep{gupta2025scalable, chen2024inside}. 

Motivated by these limitations, we propose~\method, a simple test-time scaling approach that uses hidden representations from the penultimate layer as a quality signal for candidate selection. \method~trains a lightweight scoring model on model-generated responses labeled by objective correctness, typically using a few hundred calibration problems and their corresponding rollout-level labels, making per-model training cheap while enabling model-specific calibration. Using the last non-padding token embedding from the penultimate layer to rank candidates, \method~enables Best-of-$N$ selection that is calibrated to the target model, with minimal overhead. Across the OLMo-2, Qwen3, and GPT-OSS model families, spanning 1.7B to 30B parameters, \method~outperforms prior confidence-based methods by up to $9\%$ on code generation and mathematical reasoning tasks, and matches stronger scoring models such as process reward models (PRMs), which are trained to score intermediate reasoning steps, while using up to 700$\times$ fewer parameters. To summarize, our main contributions are:

\begin{mainbox}{}
    \begin{enumerate}
    \item We demonstrate through comprehensive experiments that existing model uncertainty metrics based on logits or token probabilities yield limited performance gains for test-time scaling. (\autoref{fig:metric_selection})
    \item We propose~\method{}, a simple yet effective test-time scaling method that learns a scoring model on a small calibration dataset, enabling model-and-domain-specific adaptation. (\autoref{sec:method})
    \item Extensive experiments across three model families (OLMo-2, Qwen3, GPT-OSS) and scales (1.7B–30B) on code generation and mathematical reasoning tasks show that \method{} achieves improvements of up to $\mathbf{\sim9\%}$ over confidence-based methods and up to $\mathbf{\sim7\%}$ over a strong PRM baseline, while being over $1000\times$ faster and requiring up to $700\times$ fewer parameters. (\autoref{sec:results}).
    \end{enumerate}
\end{mainbox}
\section{Related Work}

Recent work on test-time scaling \citep{snell2024scalingllmtesttimecompute, brown2024largelanguagemonkeysscaling, wu2025inferencescalinglawsempirical} shows that additional inference-time compute can substantially improve LLM performance without further training. A central challenge that remains unsolved is how to select among multiple candidate responses. Best-of-$N$ (BoN) typically uses fixed confidence or uncertainty heuristics derived from token probabilities \citep{fu2025deepthinkconfidence, fadeeva2024factcheckingoutputlargelanguage, kang2025scalablebestofnselectionlarge}, but such signals can be unreliable across tasks and often fail to capture the semantic quality of response-level (see \ref{sec:metric_experiments_results}). Alternatively, self-consistency \citep{wang2023selfconsistencyimproveschainthought} relies on majority voting to aggregate candidates. Although highly effective for tasks with a single objective answer, this method struggles in open-ended domains such as code generation, where several valid solutions can exist. Additionally, it disregards the valuable information embedded in the intermediate reasoning steps. Finally, scaling up the number of candidates yields diminishing returns for both methods \citep{huang2025bestofnbestthemcoverage}. Other approaches use LLMs as selectors, prompting models to evaluate or rank candidates \citep{chen2023universalselfconsistencylargelanguage, xiong2024llmsexpressuncertaintyempirical, ren2023selfevaluationimprovesselectivegeneration, huang2025efficienttesttimescalingselfcalibration}. However, this approach increases inference cost and risks inheriting biases from the selector. More broadly, recent approaches aggregate multiple responses through learned aggregation, recursive refinement, or trained scoring models \citep{li2025llmsgeneratebetteranswer, wang2025learningrefineselfrefinementparallel, zhao2025majorityrightrltraining, venkatraman2025recursiveselfaggregationunlocksdeep, cobbe2021trainingverifierssolvemath}. These include outcome reward models that score final answers \citep{uesato2022solvingmathwordproblems, cobbe2021trainingverifierssolvemath}, and process reward models that score intermediate reasoning steps \citep{lightman2023letsverifystepstep, wang2024mathshepherdverifyreinforcellms, li2025processrewardmodelqvalue, snell2024scalingllmtesttimecompute}. While effective, such methods typically require additional supervision and expensive scoring models. In contrast, our approach is lightweight, requires minimal additional data and uses only a lightweight scoring head.

\section{Background: Best-of-$N$ Selection}
\label{background}

Test-time scaling improves output quality by using additional inference-time compute. In the parallel setting, a language model samples multiple candidate rollouts $\{y_1,\dots,y_N\}$ for a prompt $x$, and then selects or aggregates them. A common strategy is \emph{Best-of-$N$}, which returns the rollout with the highest score under a scoring function $f(\cdot)$: $y^\star = \arg\max_{j=1,\dots,N} f(y_j)$. To define the scoring function $f$, recent methods~\citep{fu2025deepthinkconfidence, kang2025scalablebestofnselectionlarge} derive confidence metrics from token-level log probabilities to rank candidate responses. A common approach is confidence-based scoring, which uses model probability signals~\citep{fu2025deepthinkconfidence, kang2025scalablebestofnselectionlarge}. Concretely, token-level confidence at position $i$ is defined as $C_i = -\frac{1}{k} \sum_{j=1}^{k} \log P_i(j)$, where $P_i(j)$ is the probability of the $j$-th most likely token. A sequence-level score is then obtained by aggregating these values over a selected set of token positions: $f(y;\mathcal{I}) = \frac{1}{|\mathcal{I}(y)|} \sum_{i \in \mathcal{I}(y)} C_i$, where $\mathcal{I}(y)$ denotes the included indices. Existing metrics differ mainly in the choice of $\mathcal{I}(y)$, for example using all tokens, a suffix of the sequence, or low-confidence token groups. We summarize several popular variants in~\ref{app:confidencebasedscoring}, define additional probability-based scoring functions in~\ref{app:varuncertainity}, and defer a broader discussion to~\autoref{sec:metric_experiments_results}. As shown in~\autoref{fig:metric_selection}, uncertainty-based metrics often fail to reliably identify correct reasoning traces and frequently perform near random selection. This limitation likely stems from poor signal calibration; models are often over-confident in incorrect responses and under-confident in correct ones, which limits their reliability for selection~\citep{epstein2025llms}. This motivates a different selection signal for Best-of-N. While prior uncertainty-based metrics~\citep{fu2025deepthinkconfidence, chen2023universalselfconsistencylargelanguage} succeed in structured settings that utilize majority voting, raw logits and token probabilities lack discriminatory signals for reliable BoN selection in free-form tasks such as code generation, where agreement-based aggregation is not well defined.

\section{Our Method: \method}
\label{sec:method}

\begin{figure}[t]
    \centering
    \includegraphics[width=1\linewidth]{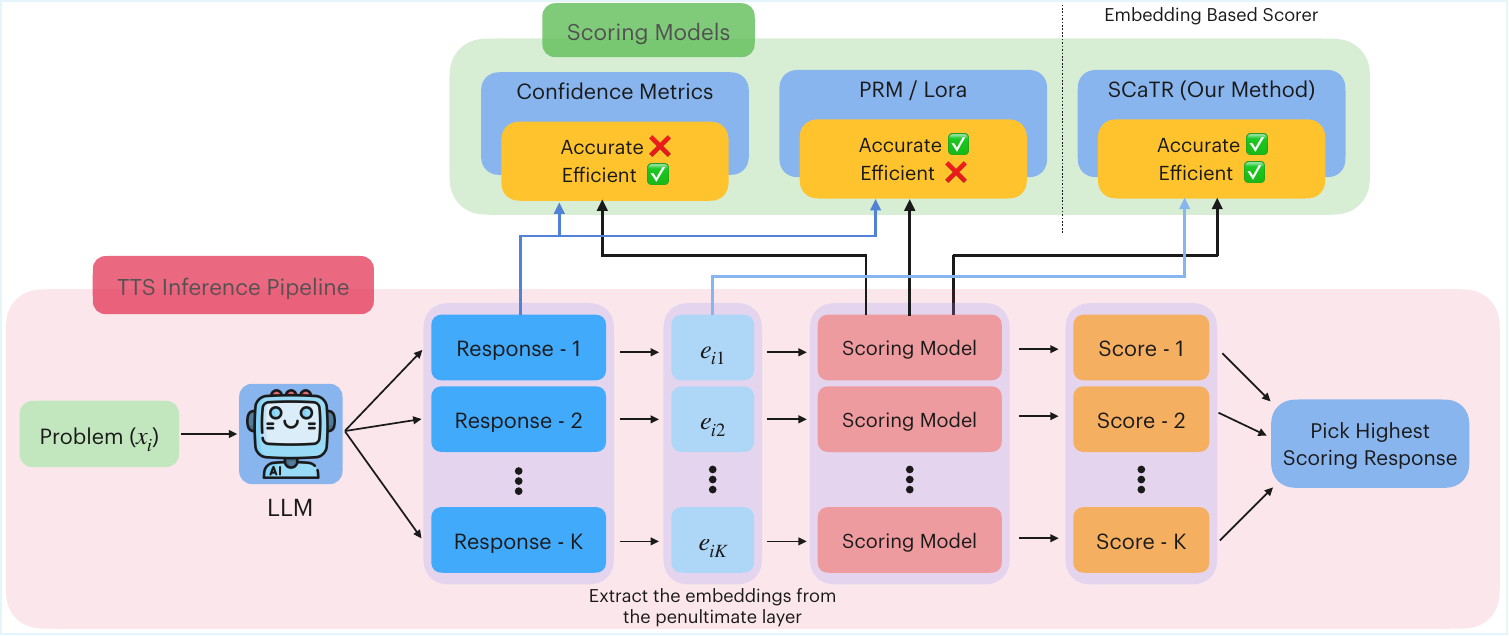}
    \caption{Overview of \method. Given an input, the model generates $K$ candidate responses. For each response, we extract the intermediate embedding of the last non-padding token from the penultimate layer. These embeddings are evaluated by a scoring model trained on a calibration set, and the response with the highest score is selected as the final output.}
    \vspace{-2mm}
    \label{fig:placeholder}
\end{figure}

Motivated by recent work showing that internal model states can support downstream decision-making~\citep{chen2024inside, gupta2025scalable}, we propose~\method, which uses the model's internal representations together with a lightweight learned scorer to guide BoN selection. \method~operates in two phases: calibration and evaluation. In the calibration phase, a small set of rollouts labeled for correctness is used to train a model-specific scoring function. Then, in the inference / evaluation phase, the LLM generates $N$ candidate responses for each new prompt; the scoring function evaluates these candidates, and the highest-scoring response is selected as the final output.

\subsection{Calibration Phase}

In the calibration phase, \method{} trains a lightweight MLP scorer on a small labeled set of prompt-response pairs with binary labels, which we refer to as the calibration set. To predict whether a response is correct, the scorer takes as input the hidden representation of the final non-padding token from the penultimate layer of the model, extracted during generation. These representations reflect the model's state after processing the full response, capturing contextual signals that are not directly available from token-level confidence metrics. We train a separate scorer for each model $M$. Below, we formalize the calibration set construction and the learning procedure for the scoring function.

\paragraph{Calibration set.} To learn the scoring function, we construct a \textit{calibration set} that captures the relationship between the internal states of the model and the correctness of its outputs. Let the original dataset of reasoning tasks (i.e. prompts) be $\{x_i\}_{i=1}^m.$
For each task \(x_i\), we generate \(N\) candidate responses (i.e. rollouts) using the language model \(M\), $y_{ij} \sim M(\cdot \mid x_i),\ \forall \ j \in \{1,\ \dots,\  N\}$. We denote $\mathcal{X}$ as the space of prompts and $\mathcal{Y}$ as the space of outputs. Note that each response is evaluated for correctness using an evaluator 
\(\mathcal{E} : \mathcal{X} \times \mathcal{Y} \to \{0,1\}\), yielding labels $
r_{ij} = \mathcal{E}(x_i, y_{ij})$.
The evaluator is domain-specific. For example, \(r_{ij}=1\) if the model's output matches the exact solution for mathematical reasoning tasks or passes all execution tests for code generation, and \(r_{ij}=0\) otherwise. To leverage the internal computations of the model, we extract the hidden representation $\bm{e}_{ij}$ of the final non-padding token at the penultimate layer for each response $j$ of every prompt $i$. The embedding of this final non-padding token captures the model’s aggregated state after processing the entire sequence. The resulting calibration set is defined as $
\mathcal{D}_{\mathrm{cal}} = \big\{(\bm{e}_{ij}, r_{ij}) \mid \ \forall \ i \in \{1,\  \dots,\  m,\} \; \forall \ j \in \{ 1,\  \dots,\  N\} \big\}$, which provides paired examples of internal response representations and correctness labels.

\paragraph{Learning the scoring function.}  
Using \(\mathcal{D}_{\mathrm{cal}}\), we train a model-specific scoring function using the extracted embeddings from the penultimate layer: $
f_\theta : \mathbb{R}^d \to [0,1]$, parameterized as a shallow $l$-layer MLP:
$f_\theta(\bm{e}) = \sigma\Big(\bm{W}^{(l)} \phi(\cdots \phi(\bm{W}^{(1)} \bm{e}))\Big), \quad \theta = \{\bm{W}^{(k)}\}_{k=1}^l$  
where \(\phi\) is a nonlinear activation (ReLU) and \(\sigma\) is a sigmoid function. The parameters \(\theta\) are optimized using a weighted cross-entropy loss: $\mathcal{L}(\bm{x},y) = - w \cdot y \log f_\theta(\bm{x}) - (1-y) \log (1 - f_\theta(\bm{x}))$,
where \(w = N_0 / N_1\) is the ratio of negative to positive training samples. Hyperparameter details can be found in App.~\ref{sec:hyps}. For \method~, we use embeddings from the penultimate layer. We ablate on this layer choice: results can be found in App.~\ref{sec:layer}.
By training on this calibration set, the MLP learns to map hidden states of the LLM to an estimate of correctness.

\subsection{Inference / Evaluation Phase}

During evaluation, given a new prompt $x_i$, the LLM generates $N$ candidate responses ${y_1,\dots,y_N}$ in parallel. For each response $y_j$, we extract the hidden representation $\bm{e}_{ij}$ of the final non-padding token from the penultimate layer and compute its score using the learned scoring function. The final prediction is the candidate with the highest score. This is extracted during generation, implying that no additional overhead is incurred in this process. These embeddings are then passed through the learned scoring function to obtain response scores $
s_{ij} = f_\theta^{(M,D)}(\bm{e}_{ij}) \in [0,1], \ \forall  j \in \{1,\ \dots,\ N\}$, which provide an estimate of the correctness or quality of each candidate response. For free form text generation tasks such as coding tasks, the response with the highest score is selected (Best-of-\(N\)): $y^\star = \arg\max_{j=1,\dots,N} s_j$. For tasks with a definite single correct answer, such as math tasks, we additionally apply weighted majority vote as an alternative selection strategy. For each candidate answer \(a \in \mathcal{A}\), we compute the score $S(a) = \sum_{j=1}^{N} s_j \mathbf{1}\{y_j = a\}$, and select an answer \(a^\star = \arg\max_{a \in \mathcal{A}} S(a)\). In our experiments, we train \(f_\theta^{(M,D)}\) on one dataset and evaluate it on a different dataset from the same domain in order to assess the generalization of the calibration function. We consider this cross-dataset transfer setting more indicative of practical utility, since in realistic deployments the scorer must generalize to new but related questions rather than only to the specific dataset on which it was trained.
\section{Experiments} 
\label{sec:experiments}

We evaluate \method{} along three axes: effectiveness, efficiency, and robustness. Specifically, we ask: \textbf{(RQ1)} does \method{} improve Best-of-$N$ selection over lightweight confidence heuristics, \textbf{(RQ2)} how does it compare with stronger learned verifiers such as PRMs and LoRA-based selectors, and \textbf{(RQ3)} which design choices matter most in practice? To answer these questions, we study coding and mathematical reasoning tasks, report both accuracy and efficiency, and perform ablations on the representation, calibration budget, and scoring architecture. \looseness=-1

\paragraph{Datasets.}
We evaluate on both code generation and mathematical reasoning. For coding, our main results use HumanEval~\citep{chen2021evaluatinglargelanguagemodels} and a 1K-problem subset of KodCode~\citep{xu2025kodcodediversechallengingverifiable}, following \citet{ekbote2025murphy}. For mathematics, we use AIME24, AIME25, and MATH-500~\citep{lightman2023letsverifystepstep}.

\paragraph{Models.}
We study five reasoning LLMs spanning a broad range of scales and model families: Qwen3-1.7B~\citep{qwen3technicalreport}, OLMo-2-1124-7B-Instruct~\citep{olmo20252olmo2furious}, Qwen2.5-14B-Instruct~\citep{qwen2025qwen25technicalreport}, GPT-OSS-20B~\citep{openai2025gptoss120bgptoss20bmodel}, and Qwen3-30B-A3B~\citep{qwen3technicalreport}. Full model and implementation details, including hidden-state dimensionality and layer counts, are deferred to Appendix~\ref{app:details}.

\paragraph{Implementation details.}

Unless otherwise stated, we generate $N=16$ candidate rollouts per prompt. For each model--dataset pair, we construct three different calibration sets by resampling prompts and train a separate scorer on each one. In practice, \method{} requires only a single calibration set to operate; using three such sets here allows us to measure variability across calibration data choices. For coding tasks, we assign binary labels of $1$ if the generated program passes all test cases and $0$ otherwise; for math tasks, we assign $1$ if the final solution is correct and $0$ otherwise. From every rollout, we extract the final non-padding token embedding from the \emph{penultimate} transformer layer, which serves as the input to \method. We use a held-out validation split for early stopping and report mean and standard deviation across the three calibration sets. This setup intentionally keeps the scorer lightweight, but it also means that \method{} is calibrated separately for each model.

\paragraph{Baselines.}
We compare \method{} against random selection, majority voting for math tasks, and several confidence-based ranking rules derived from token probabilities: Average Trace Confidence ($C_{\mathrm{avg}}$), Bottom-10\% Group Confidence ($C_{\mathrm{bottom\text{-}10\%}}$), Lowest Group Confidence ($C_{\mathrm{least}}$), and Tail Confidence ($C_{\mathrm{tail}}$) (See \autoref{app:confidence_definitions_and_metrics} for more details). We also compare against learned verifiers: ReasonFlux-1.5B~\citep{zou2025reasonfluxprmtrajectoryawareprmslong}, and a LoRA-based correctness predictor obtained by fine-tuning the same base model used for rollout generation. Details of all baselines, including tail/window sizes and the LoRA setup, are provided in \autoref{app:details}.

\subsection{Main Results}
\label{sec:results}

\begin{figure}[t]
    \centering
    \includegraphics[width=\linewidth]{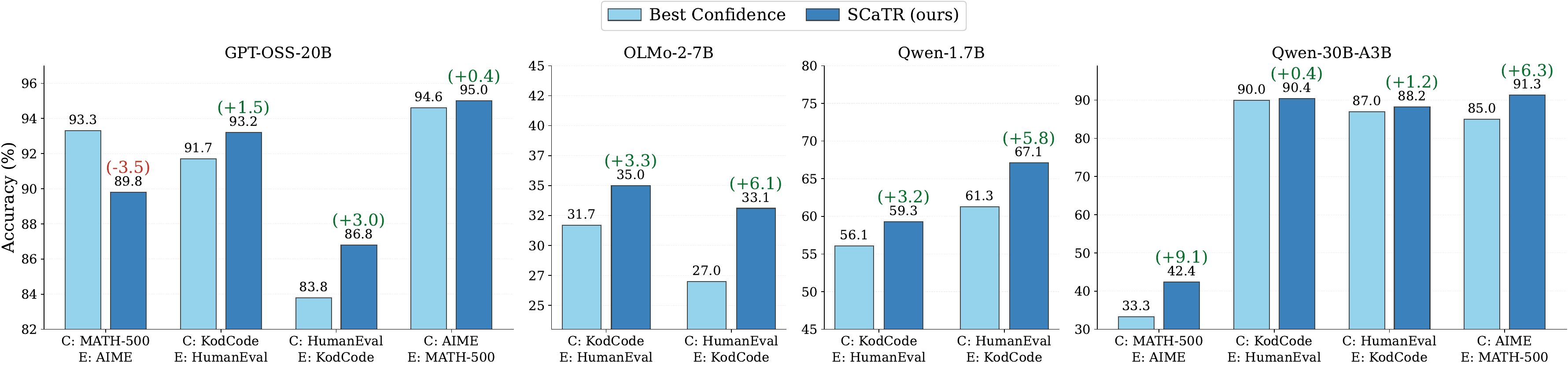}
    \caption{Comparison of \method{} with the strongest confidence-based baseline. The $x$-axis indicates the calibration dataset (C) and evaluation dataset (E). \method{} consistently matches or improves upon confidence-based selection, with gains of up to 9.1 points on mathematical reasoning and 6.1 points on coding. }
    \vspace{-2mm}
    \label{fig:main_results_all_models}
\end{figure}

\paragraph{RQ1: Does \method{} outperform lightweight confidence heuristics?}
\autoref{fig:main_results_all_models} compares \method{} with the strongest confidence-based baseline for each setting. Overall, \method{} consistently matches or improves upon confidence-based selection, with the largest gains reaching $9.1\%$ on math and $6.1\% $ on coding. For example, on Qwen-30B-A3B, \method{} improves over the best confidence heuristic by $9.1\%$ and $6.3\%$ points on AIME and Math-500 datasets. On OLMo-2-7B, the gains are $3.3\%$ on HumanEval and $6.1\%$ on KodCode. Across settings, \method{} also bridges a substantial fraction of the gap between the strongest heuristic baseline and oracle selection. These results indicate that fixed confidence scores capture useful but limited information, whereas a lightweight learned scorer can exploit richer signals from hidden representations.

For mathematical reasoning, we additionally combine \method{} with weighted majority voting (WMV). This further improves performance in several settings, suggesting that \method{} is complementary to answer aggregation when a canonical final answer is available in~\autoref{tab:main_results_math}.

\paragraph{RQ2: How does \method{} compare with stronger learned verifiers?}
We next compare \method{} with learned verifiers, focusing on both accuracy and latency cost.

On math, \autoref{tab:scatr_vs_reasonflux} compares \method{} with ReasonFlux using Qwen2.5-14B rollouts. \method{} outperforms ReasonFlux-1.5B on all three reported benchmarks, despite being trained on only 1K OpenThoughts examples, randomly sampled as 500 math and 500 code problems, whereas ReasonFlux is trained on the full 114K-example OpenThoughts dataset. This result highlights a central advantage of \method{}: model-specific calibration can be highly data-efficient while still matching or surpassing substantially more data-intensive PRM approaches. Even relative to ReasonFlux-7B, \method{} remains competitive on several settings (see~\ref{sec:reasonfluxexp}) despite being vastly smaller and requiring far less supervision, making it a compelling lightweight alternative to large PRMs.

\begin{table}[t]
\centering
\small
\begin{tabular}{l c c c}
\toprule
\textbf{Dataset} & \textbf{ReasonFlux-1.5B} & \textbf{\method} & \textbf{$\Delta$ Accuracy $\uparrow$} \\
\midrule
AIME24   & 17.8$\pm$1.9 & \textbf{18.5$\pm$4.5} & \cellcolor{green!8}{+0.7} \\
AIME25   & 14.4$\pm$1.9 & \textbf{21.6$\pm$5.5} & \cellcolor{green!18}{+7.2} \\
MATH-500 & 79.2$\pm$0.4 & \textbf{80.4$\pm$0.7} & \cellcolor{green!10}{+1.2} \\
\bottomrule
\end{tabular}
\vspace{0.5em}
\caption{Comparison with ReasonFlux-1.5B on mathematical reasoning using Qwen2.5-14B rollouts. \method{} is trained on only 1K OpenThoughts examples (500 math and 500 code), whereas ReasonFlux is trained on the full 114K OpenThoughts dataset.}
\vspace{-2mm}
\label{tab:scatr_vs_reasonflux}
\end{table}

\begin{table*}[t]
\centering
\scriptsize
\setlength{\tabcolsep}{3pt}
\renewcommand{\arraystretch}{0.8}
\vspace{-2mm}
\begin{tabular}{ll|ccc|ccccc}
\toprule
\multirow{2}{*}{\textbf{Model}} & \multirow{2}{*}{\textbf{Dataset}} & \multicolumn{3}{c|}{\textbf{ReasonFlux-PRM-1.5B}} & \multicolumn{5}{c}{\textbf{\method~ (ours)}} \\
\cmidrule(lr){3-5} \cmidrule(lr){6-10}
 & & \textbf{Size} & \textbf{Infer (ms) $\downarrow$} & \textbf{Acc. $\uparrow$} & \textbf{Size} & \textbf{Infer (ms) $\downarrow$} & \textbf{Speedup $\uparrow$} & \textbf{Acc. $\uparrow$} & \textbf{$\Delta$ Accuracy $\uparrow$} \\
\midrule

\multirow{2}{*}{GPT-OSS-20B}
& HumanEval & 1.5B & 132.3 & 89.8$\pm$2.5 & 5.0M & 0.18 & \textcolor{darkgreen2}{717.2$\times$} & 93.2$\pm$0.6 & \cellcolor{green!16}{+3.3} \\
& KodCode   & 1.5B & 137.4 & 82.5$\pm$4.0 & 2.9M & 0.20 & \textcolor{darkgreen2}{678.6$\times$} & 86.8$\pm$0.7 & \cellcolor{green!20}{+4.2} \\

\midrule
\multirow{2}{*}{OLMo-2-7B}
& HumanEval & 1.5B & 55.3 & 32.9$\pm$1.6 & 6.6M & 0.16 & \textcolor{darkgreen2}{348.6$\times$} & 35.0$\pm$1.7 & \cellcolor{green!10}{+2.1} \\
& KodCode   & 1.5B & 54.8 & 29.3$\pm$1.3 & 2.2M & 0.20 & \textcolor{darkgreen2}{279.6$\times$} & 33.1$\pm$0.9 & \cellcolor{green!18}{+3.8} \\

\midrule
\multirow{2}{*}{Qwen-1.7B}
& HumanEval & 1.5B & 116.7 & 60.4$\pm$4.3 & 2.6M & 0.15 & \textcolor{darkgreen2}{757.7$\times$} & 59.3$\pm$8.1 & \cellcolor{red!12}{-1.1} \\
& KodCode   & 1.5B & 79.4 & 64.5$\pm$6.9 & 2.2M & 0.18 & \textcolor{darkgreen2}{441.3$\times$} & 67.1$\pm$12.2 & \cellcolor{green!12}{+2.6} \\

\midrule
\multirow{2}{*}{Qwen-30B-A3B}
& HumanEval & 1.5B & 25.2 & 87.6$\pm$0.7 & 2.7M & 0.20 & \textcolor{darkgreen2}{111.5$\times$} & 90.4$\pm$0.8 & \cellcolor{green!13}{+2.8} \\
& KodCode   & 1.5B & 38.2 & 84.6$\pm$1.1 & 2.1M & 0.20 & \textcolor{darkgreen2}{188.7$\times$} & 88.2$\pm$0.6 & \cellcolor{green!17}{+3.6} \\

\bottomrule
\end{tabular}
\vspace{1em}
\caption{Efficiency and accuracy comparison of ReasonFlux-PRM-1.5B~\citep{zou2025reasonfluxprmtrajectoryawareprmslong} and \method~ across evaluation datasets for coding tasks. SCATR achieves comparable or higher accuracy while using orders of magnitude fewer parameters and providing substantial inference speedups of up to 1000$\times$.}
\vspace{-2mm}
\label{tab:scatr_efficiency_comparison}
\end{table*}

Next, we next examine the accuracy--efficiency trade-off in \autoref{tab:scatr_efficiency_comparison} and \autoref{tab:scatr_efficiency_comparison_math_reasonflux}. Across models and domains, \method{} achieves competitive, and often higher, accuracy than PRM-based selectors while using up to $1{,}000\times$ fewer parameters and enabling up to $1{,}000\times$ faster inference. In several settings, \method{} also improves absolute accuracy, with gains of up to 7.8 points on mathematical benchmarks and 4.2 points on coding datasets. These results show that the efficiency benefits of \method{} do not come at the expense of selection quality.

We also compare \method{} against a LoRA-based correctness predictor obtained by fine-tuning the full base model. Despite its simplicity, \method{} remains competitive with, and often outperforms, LoRA across a range of settings. For example, on GPT-OSS-20B, \method{} matches LoRA on AIME (0.95 vs.\ 0.95) and improves over it on KodCode (0.93 vs.\ 0.92) and HumanEval (0.87 vs.\ 0.85). At the same time, \method{} is substantially more efficient, using up to $8{,}000\times$ fewer parameters and enabling up to $3{,}500\times$ faster inference. Additional results across more models and datasets are provided in \autoref{app:additional_exps_scatr} (\autoref{tab:main_results_math} and \autoref{tab:main_results_code}).

\begin{figure}[h]
    \centering
    \includegraphics[width=\linewidth]{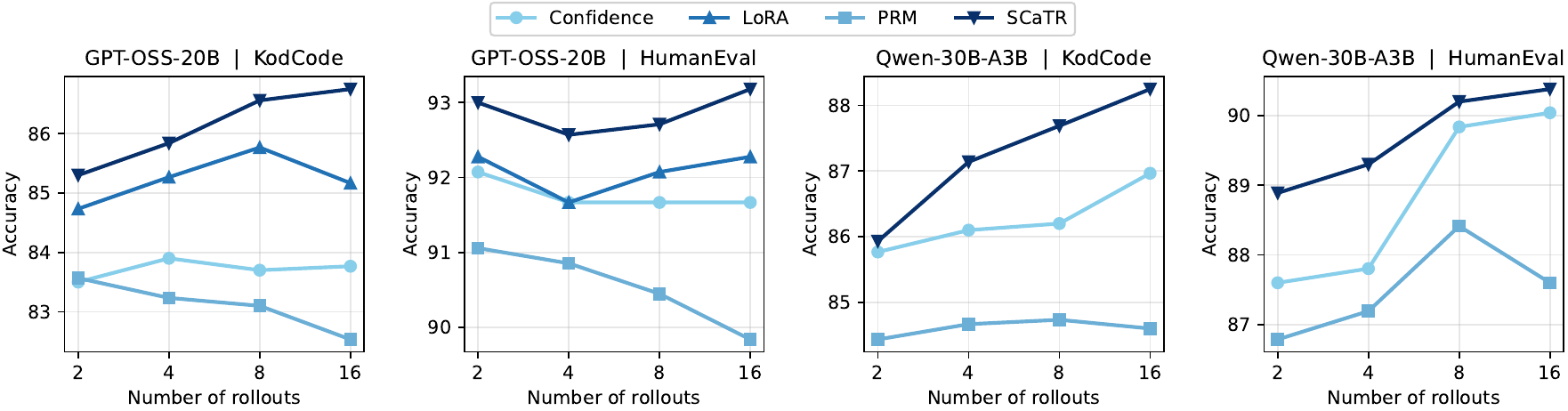}
\caption{Accuracy as a function of the number of rollouts on two models and two coding benchmarks. In each experiment, the dataset named in the figure title is the evaluation dataset, and the calibration set is taken from the other coding dataset (HumanEval or KodCode). \method{} remains competitive across rollout budgets and improves steadily as more candidates are available.}
    \vspace{-2mm}
    \label{fig:performance_num_rollouts}
\end{figure}

Finally, we further analyze performance as a function of the number of generated rollouts, as shown on~\autoref{fig:performance_num_rollouts}. \method~consistently outperforms confidence-based selection, PRM, and LoRA across all rollout budgets on coding datasets across GPT-OSS and Qwen3-30B models, exhibiting steady performance gains as the number of rollouts increases, demonstrating effective usage of additional candidate solutions for improved selection. However, these gains might not generalize to all domains. We notice plateauing or degradation of performance on some math tasks, especially for smaller and less capable models, as noted in~\autoref{app:additional_exps_scatr} (\autoref{fig:performance_num_rollouts_math}).

\subsection{Ablation Studies}
\label{sec:ablation}

\paragraph{RQ3: Which design choices matter most for \method?}
We study three factors that most directly affect the practicality and effectiveness of \method: i) the choice of hidden representation, ii) the size of the calibration set, and iii) the complexity of the scoring model.
We also study in-domain and cross-domain performance versions of \method\ in~\autoref{app:additional_experiments}.

\begin{figure}[t]
    \centering
    \vspace{-0mm}
    \includegraphics[width=\linewidth]{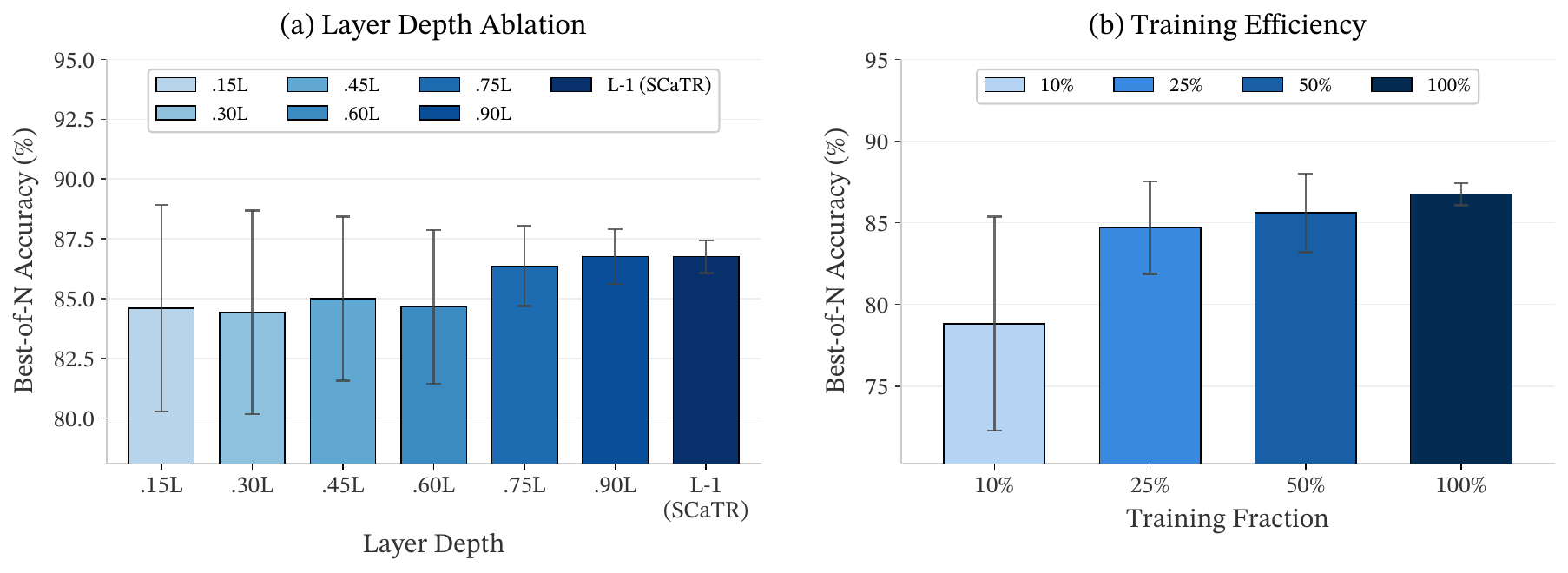}
    \caption{Illustrative ablations for GPT-OSS-20B calibrated on HumanEval and evaluated on KodCode. \textbf{Left:} effect of layer choice (total layers $L=24$) for the final-token embedding; the rightmost bar corresponds to \method. Penultimate layer exhibits robust performance with relatively low variance. \textbf{Right:} effect of reducing the calibration-set size. Larger calibration set leads to improved performance with reducing variance. Additional results are reported in Appendix~\ref{sec:layer} and Appendix~\ref{sec:fraction}. }
    \vspace{-2mm}
    \label{fig:ablations_combined}
\end{figure}

\paragraph{Choice of layer.}
Our default choice is the final non-padding token embedding from the penultimate layer, denoted $L\!-\!1$. To test this choice, we compare against six alternative layers spaced throughout the network: $.15L$, $.3L$, $.45L$, $.6L$, $.75L$, and $.9L$, where $L$ is the total number of transformer layers. In the experiment shown in~\autoref{fig:ablations_combined} (left), we use GPT-OSS-20B calibrated on HumanEval and evaluated on KodCode. For each model--dataset pair, we train a separate MLP scorer for each layer and evaluate Best-of-$N$ accuracy on the held-out test set. As illustrated in~\autoref{fig:ablations_combined} (left), performance is similar across deeper layers, and no single layer is uniformly best across all settings. The penultimate layer is therefore not uniquely optimal, but it is a robust choice that performs well across models while exhibiting relatively low variance. This makes it a good default for the main experiments.

\paragraph{Size of calibration set.}
We next vary the size of the calibration set to measure the data efficiency of \method. Using GPT-OSS-20B, we calibrate on HumanEval and evaluate on KodCode, while subsampling the HumanEval calibration set (164 problems) to 10\%, 25\%, and 50\% of the full set, and keeping $N=16$ rollouts per prompt fixed. The results in~\autoref{fig:ablations_combined} (right) show that performance degrades gracefully as the calibration budget is reduced: larger calibration sets improve performance and generally reduce variance. Because the calibration source here is HumanEval, which is already relatively small, this trend is less saturated than in corresponding GPT-OSS-20B experiments using larger calibration sources such as KodCode. On larger calibration sources such as KodCode, the gains begin to plateau before using the full dataset, suggesting that \method{} can remain effective even with substantially fewer labeled examples. Additional results are reported in Appendix~\ref{sec:fraction}.

\paragraph{Scoring function design.}
We finally compare \method~with five more complex scoring architectures: \textbf{1. single-best-layer}, which trains one MLP per candidate layer and selects the layer with the best validation loss; \textbf{2. ensemble}, which combines per-layer MLP predictions through a learned convex combination; \textbf{3. transformer-final}, which applies a Transformer encoder to the sequence of final-token embeddings across layers; \textbf{4. transformer-last-10}, which encodes the last ten token embeddings from the final layer; and \textbf{5. transformer-special}, which uses task-specific token positions such as return statements and function-definition colons in code.
We tabulate our results in~\autoref{tab:scoring_func_design}. Across these alternatives, the overall picture is clear: added architectural complexity does not deliver consistent gains. In several settings, the more complex models match \method{}, but they do not reliably exceed it, and in others they underperform despite higher training and inference cost. This supports our central design choice: a simple scorer on a single penultimate-layer representation is already a strong operating point. \looseness=-1

\begin{table}[t]
\centering
\label{tab:ablation_main}
\resizebox{\linewidth}{!}{
{
\small
\setlength{\tabcolsep}{4pt}
\begin{tabular}{lll|ccccc|cc}
\toprule
\textbf{Model} & \textbf{Train} & \textbf{Test} & \textbf{Final} & \textbf{Last-10} & \textbf{Special} & \textbf{Ensemble} & \textbf{Best Layer} & \textbf{\method\ $\uparrow$} & \textbf{$\Delta$ Accuracy $\uparrow$} \\
\midrule
Qwen-30B-A3B & HumanEval & KodCode   & 88.0 $\pm$ 0.8 & 86.7 $\pm$ 1.2 & 87.7 $\pm$ 0.7 & 87.8 $\pm$ 0.7 & 86.8 $\pm$ 1.3 & \textbf{88.3 $\pm$ 0.6} & \cellcolor{green!8}{+0.3} \\
Qwen-30B-A3B & KodCode   & HumanEval & 90.8 $\pm$ 0.7 & \textbf{90.9 $\pm$ 0.6} & 90.8 $\pm$ 0.8 & 90.7 $\pm$ 0.7 & \textbf{90.9 $\pm$ 0.8} & 90.4 $\pm$ 0.8 & \cellcolor{red!8}{-0.5} \\
\midrule
GPT-OSS-20B  & HumanEval & KodCode   & 86.5 $\pm$ 1.2 & 85.7 $\pm$ 1.0 & 85.5 $\pm$ 2.0 & 86.4 $\pm$ 2.4 & 85.2 $\pm$ 2.9 & \textbf{86.8 $\pm$ 0.7} & \cellcolor{green!8}{+0.3} \\
GPT-OSS-20B  & KodCode   & HumanEval & 93.1 $\pm$ 0.8 & 92.2 $\pm$ 0.8 & 92.4 $\pm$ 0.7 & \textbf{93.3 $\pm$ 0.5} & 93.2 $\pm$ 0.6 & 93.2 $\pm$ 0.6 & \cellcolor{red!8}{-0.1} \\
\midrule
OLMo-2-7B    & HumanEval & KodCode   & 33.0 $\pm$ 1.1 & 30.2 $\pm$ 0.8 & 29.2 $\pm$ 1.3 & \textbf{34.0 $\pm$ 1.0} & 32.7 $\pm$ 1.2 & 33.1 $\pm$ 0.9 & \cellcolor{red!8}{-0.9} \\
OLMo-2-7B    & KodCode   & HumanEval & 34.9 $\pm$ 2.3 & 33.9 $\pm$ 2.5 & 31.0 $\pm$ 1.8 & \textbf{35.2 $\pm$ 1.7} & 34.8 $\pm$ 3.4 & 35.0 $\pm$ 1.7 & \cellcolor{red!8}{-0.2} \\
\bottomrule
\end{tabular}
}
}
\vspace{1em}
\caption{We compare five more sophisticated scoring function design mechanisms to \method. Some are transformer-based, while others use intermediate layers instead of the penultimate layer like \method. We see that results are very similar across methods, all while \method~takes a fraction of the training cost. In particular, \method~shows the highest mean or is always within standard deviation of the highest mean among other methods.}
\vspace{-2mm}
\label{tab:scoring_func_design}
\end{table}
\section{Conclusion \& Limitations}

We presented \method{}, a simple and efficient test-time ranking method for Best-of-$N$ selection that leverages intermediate representations instead of token-level confidence signals. Across models and tasks in code generation and mathematical reasoning, \method{} consistently improves over confidence-based baselines and remains competitive with stronger learned verifiers while offering a favorable accuracy--efficiency trade-off. These results highlight the value of internal model states for lightweight, calibrated selection at inference time. At the same time, \method{} requires access to intermediate hidden representations, which restricts its applicability to only open-weight models. Our evaluation focuses on single-turn reasoning tasks; extending the method to multi-turn settings, where internal states evolve across interactions, remains an open direction. In addition, we restrict the scorer to a single summary representation, the final non-padding token embedding, and do not explore richer internal signals such as token-level states, attention patterns, or layer-wise dynamics, which may further improve performance on longer or more complex reasoning problems.

\newpage

\bibliography{colm2026_conference}
\bibliographystyle{colm2026_conference}

\newpage

\appendix

\appendix

\vspace{1em}
\begin{center}
\rule{\textwidth}{1pt}\\[0.75em]
{\LARGE\bfseries Appendix}\\[0.5em]
\rule{\textwidth}{1pt}
\end{center}
\vspace{1em}
\tableofcontents

\newpage

\section*{Appendix organization}
The appendix is organized into three sections covering experimental setup details, definitions and metrics, and additional experimental results including ablations and extended analyses.  \autoref{app:details} specifies the experimental framework, including model architectures and hidden-state dimensionalities, dataset compositions and partitions, the full set of baseline configurations and hyperparameter details for \method. \autoref{app:confidence_definitions_and_metrics} provides an overview of existing confidence-based metrics and formalizes the evaluated token-level uncertainty metrics (e.g. variance, Shannon entropy, probability gap) and trace-level aggregation strategies. 
Finally, \autoref{app:additional_results_full_section} presents comprehensive additional experiments. We first report extended results for SCATR in \autoref{app:additional_exps_scatr}, including Best-of-$N$ evaluations across mathematical reasoning and code generation benchmarks, demonstrating that our lightweight scoring formulation achieves performance competitive with compute-intensive baselines such as LoRA fine-tuning and ReasonFlux PRM, while requiring orders-of-magnitude fewer training FLOPs and delivering up to $1000\times$ inference speedups. We then present additional results for confidence-based methods in \autoref{sec:metric_experiments_results}, showing that standard confidence-based heuristics fail to provide robust discriminative signals for reasoning trajectory selection, performing close to random sampling. Next, we include ablation studies in \autoref{app:extended_ablations_section} analyzing the impact of key design choices, including layer selection, calibration set size, and scoring function design. Finally, we present additional experiments in \autoref{app:additional_experiments}, covering LoRA-based domain transfer, comparison with ReasonFlux PRMs, domain-adaptive inference, and SCATR performance under in-distribution and out-of-distribution settings.

\section{Experimental Setup Details}
\label{app:details}

\paragraph{Model Configurations.}
Table~\ref{tab:model_configs} reports key model configuration details, including the dimensionality of hidden representations extracted from the underlying language models. These representations are used as inputs to the scoring model in \method. We also include relevant architectural details for each model family.
For GPT-OSS-20B, we configure the reasoning effort to high during inference. All models use their default sampling temperatures. 

\paragraph{Datasets.}
Table~\ref{tab:datasets} summarizes all datasets used in our experiments, including their sizes and corresponding domains (coding or mathematical reasoning). Unless specified, calibration and evaluation are performed across paired datasets within each domain: MATH-500 and AIME for mathematical reasoning, and HumanEval and KodCode for coding, with calibration and test roles swapped between the two.

\paragraph{Baselines.}
 We compare ~\method~ against random selection, majority voting (for math tasks), and a set of confidence-based scoring methods, including Average Trace Confidence ($C_{\mathrm{avg}}$), Bottom 10\% Group Confidence ($C_{\mathrm{bottom-10\%}}$), Lowest Group Confidence ($C_{\mathrm{least}}$), and Tail Confidence ($C_{\mathrm{tail}}$), as well as PRMs, including ReasonFlux-1.5B and ReasonFlux-7B, and a LoRA-based correctness predictor obtained by fine-tuning the base model used for rollout generation. 
All the confidence-based metrics are computed using top-$k$ token probabilities ($k=10$), with a group window of 1024 tokens and a tail length of 2048 tokens. For the LoRA-based correctness predictor, we fine-tune low-rank adapters on the base rollout model using separate datasets. During inference, the final $k$ tokens of each response are fed to a binary yes/no classifier, and the rollout with the highest probability of “yes” is selected. We use $k=16{,}384$ tokens for all models except GPT-OSS, where $k=4{,}096$. In all cases, calibration data is drawn from a dataset distinct from the evaluation set.

\paragraph{Hyperparameter Details for \method}\label{sec:hyps}
In this section, we list the architecture, loss, and hyperparameter details for \method. The scoring function is a lightweight MLP trained on the calibration set with binary cross-entropy loss (with logits), using a class-balanced positive weight of $n_{\text{neg}}/n_{\text{pos}}$ to account for label imbalance. The architecture consists of an input dropout layer followed by 2-3 fully-connected hidden layers with ReLU activations and dropout, and a single scalar output logit; hidden dimensions are selected from $\{[512, 256], [512, 256, 128], [1024, 512], [1024, 512, 256]\}$. The training set is split 75/25 at the problem-ID level to prevent leakage across rollouts of the same problem. We use Adam with gradient clipping (norm 1.0) and a ReduceLROnPlateau scheduler (factor 0.5, patience 3), with early stopping at patience 10 per configuration and a maximum of 100 epochs. The hyperparameters: hidden dimensions, dropout, input dropout, learning rate, weight decay, batch size, and whether to use batch normalization; are selected via random search over 100 configurations, choosing the configuration with the lowest validation loss. All experiments are run with 3 random seeds (32, 42, 52) across 3 training turns (three calibration sets), and results are averaged over the resulting 9 runs.

\begin{table}[]
\centering
\small

\begin{minipage}[t]{0.48\columnwidth}
\centering
\begin{tabular}{lrr}
\toprule
\textbf{Model} & \textbf{Layers} & \thead{\textbf{Hidden}\\ \textbf{dimension}} \\
\midrule
GPT-OSS-20B          & 24 & 2880 \\
OLMo-2-7B            & 32 & 4096 \\
Qwen3-1.7B           & 28 & 2048 \\
Qwen3-30B-A3B        & 48 & 2048 \\
Qwen2.5-14B-Instruct & 48 & 5120 \\
\bottomrule
\end{tabular}
\vspace{1em}
\caption{Model architecture specifications. The hidden dimension is the feature vector size used to train the scoring function.}
\label{tab:model_configs}
\end{minipage}
\hfill
\begin{minipage}[t]{0.48\columnwidth}
\centering
\begin{tabular}{llr}
\toprule
\textbf{Dataset} & \textbf{Domain} & \thead{\textbf{Number of} \\ \textbf{problems}} \\
\midrule
MATH-500         & Math & 500  \\
AIME 2025        & Math & 30   \\
AIME 2024        & Math & 30   \\
HumanEval        & Code & 164  \\
KodCode (subsetted) & Code & 1000 \\
\bottomrule
\end{tabular}
\vspace{1em}
\caption{Dataset specifications including problem domain and number of problem in each dataset.}
\label{tab:datasets}
\end{minipage}
\vspace{-2mm}
\end{table}

\section{Definitions and Metrics}
\label{app:confidence_definitions_and_metrics}
\subsection{Confidence-based Metrics}
\label{app:confidencebasedscoring}

When the majority vote is ill-defined (e.g., multiple distinct outputs can be correct) or when correctness is difficult to verify automatically, selection must rely on signals other than exact agreement on the final answer. A common alternative is confidence-based scoring, which ranks candidates using model-internal probability signals. In the literature~\citep{fu2025deepthinkconfidence, kang2025scalablebestofnselectionlarge}, a common approach is to turn per-token log probabilities into a single confidence-style score, which is then used to rank and select among candidate generations. To make these confidence-based metrics precise, we first define a token-level confidence signal and then show how it can be aggregated into a sequence-level score. 

We begin with token-level confidence, defined as the negative average log-probability of the top-$k$ predicted tokens at position $i$: $C_i = -\frac{1}{k} \sum_{j=1}^{k} \log P_i(j)$, where $P_i(j)$ is the probability assigned to the $j$-th most likely token. Higher token confidence corresponds to more peaked predictive distributions (greater model certainty), whereas lower confidence indicates flatter distributions (less certainty) at that position.

Confidence-based scoring then squashes these token-level confidence values to a single score for generated text by aggregating them over a chosen subset of token positions (or groups of positions). All confidence-based scoring functions of a generated response can be written as a normalized aggregation of token-level confidence values over a selectable index set: $f(y;\mathcal{I}) = \frac{1}{|\mathcal{I}(y)|} \sum_{i \in \mathcal{I}(y)} C_i$, where $\mathcal{I}(y) \subseteq \{1,\ldots, T\}$ denotes the subset of token indices included in the aggregation. \emph{Average trace confidence} is obtained by selecting all token indices, $\mathcal{I}(y)=\{1,\ldots,T\}$, yielding a global measure of model certainty. To capture local uncertainty, tokens are organized into overlapping groups $\{G_j\}$, with \emph{group confidence} defined as the mean token confidence within each group. \emph{Bottom $10\%$ group confidence} is computed by identifying the lowest-confidence $10\%$ of groups and aggregating the token confidences contained in those groups, while \emph{lowest group confidence} considers only the tokens belonging to the single minimum-confidence group. \emph{Tail confidence} is obtained by restricting $\mathcal{I}(y)$ to $T_{\mathrm{tail}}$, a fixed-length suffix of the generated sequence, emphasizing uncertainty in the final part of the trace.\looseness=-1 

In prior work~\citep{fu2025deepthinkconfidence, kang2025scalablebestofnselectionlarge}, token-level confidence is typically defined as the mean of the top-$k$ log-probabilities, $C_i = \frac{1}{k} \sum_{j=1}^{k} \ell_j$, where $\ell_1,\dots,\ell_k$ are the top-$k$ log-probabilities at token position $i$. While simple and widely used, this mean-based confidence can be overly sensitive to outliers and may fail to capture the full uncertainty in the model’s predictions.

In this work, we study several alternative token-level uncertainty metrics derived from the same top-$k$ log-probabilities. 

\subsection{Variants of Uncertainty Metrics}
\label{app:varuncertainity}

We analyzed a family of probability-based, token-level scoring signals that resemble confidence and can be aggregated into a single rollout-level metric for Best-of-$N$ selection. While confidence-based scoring in prior work is typically defined using a token-level confidence $C_i$ (often computed as the mean negative log-probability of the top-$k$ tokens), we go beyond this standard formulation. In addition to the standard confidence metric, we study several alternative token-level measures derived from the model’s token-wise probabilities or log-probabilities.

We consider the median of the top-$k$ log-probabilities, which is less sensitive to outliers than the mean and can provide a more robust proxy for confidence. We also evaluate the variance of the top-$k$ log-probabilities, which captures dispersion among likely tokens and serves as a proxy for predictive uncertainty. To measure the sharpness of the model’s preference, we compute the probability gap between the two most likely tokens. Finally, we consider the Shannon entropy of the normalized top-$k$ distribution, which quantifies uncertainty in the predictive distribution. A detailed description of all scoring metrics is given below.

\paragraph{Median.} 
The \textit{median} of the top-$k$ log-probabilities provides a more robust estimate of confidence by reducing the influence of extreme values and better reflecting the central tendency of the model’s predicted probabilities. Compared to the mean, it is less sensitive to outliers and better captures the typical confidence of likely tokens.

\paragraph{Variance.} 
The \textit{variance} of the top-$k$ log-probabilities quantifies the dispersion among the top-$k$ tokens:
\[
\text{Var} = \frac{1}{k} \sum_{i=1}^{k} (\ell_i - \bar{\ell})^2, \quad \bar{\ell} = \frac{1}{k} \sum_{i=1}^{k} \ell_i.
\]
Higher variance indicates greater disagreement among likely tokens, serving as a proxy for predictive uncertainty.

\paragraph{Probability Gap.} 
To capture the sharpness of the model’s preference, we compute the \textit{probability gap}, defined as the difference between the top two probabilities:
\[
\text{Gap} = \exp(\ell_1) - \exp(\ell_2).
\]
A larger gap indicates a more confident prediction at that token position.

\paragraph{Shannon Entropy.} 
Finally, we consider the \textit{Shannon entropy} $H$, which captures the overall uncertainty of the model’s predictions at a token. It is computed over the normalized top-$k$ probabilities:
\[
H = -\sum_{i=1}^{k} \tilde{p}_i \log \tilde{p}_i, \quad \tilde{p}_i = \frac{\exp(\ell_i)}{\sum_{j=1}^{k} \exp(\ell_j)}, \quad i = 1,\dots,k.
\]
Higher entropy corresponds to a more uncertain or flat distribution over likely tokens.

Trace-level scores are computed using the group-level confidence measures introduced in \autoref{app:confidencebasedscoring}: Average Trace Confidence ($C_{\mathrm{avg}}$), Bottom 10\% Group Confidence ($C_{\mathrm{bottom-10\%}}$), Lowest Group Confidence ($C_{\mathrm{least}}$), and Tail Confidence ($C_{\mathrm{tail}}$). In our experiments, we use $k=10$ for top-$k$ token confidence, a group size of 1024 tokens, and a tail length of 2048 tokens.

For Best-of-$N$ selection, the candidate with the highest aggregate score is chosen for all metrics except entropy, for which lower values indicate higher confidence. Selection is applied across all variants of trace-level aggregation ($C_{\mathrm{avg}}$, $C_{\mathrm{bottom-10\%}}$, $C_{\mathrm{least}}$ and $C_{\mathrm{tail}}$).


\section{Additional Experimental Results}
\label{app:additional_results_full_section}

\subsection{\method~: Additional Results for ~\autoref{sec:results}}
\label{app:additional_exps_scatr}
We present full results for the math and coding domains, comparing \method~to several baseline methods and reporting the improvement over the strongest baseline. We also report the percentage of the gap between the best baseline and the oracle that is closed by \method, computed as $\frac{\text{\method} - \text{best baseline}}{\text{Oracle} - \text{best baseline}}$. For math tasks, we include both Best-of-$N$ results and Weighted Majority Vote results. Detailed results of math and coding tasks can be found in \autoref{tab:main_results_math} and \autoref{tab:main_results_code}, respectively. We show that \method~ consistently outperforms confidence-based baselines, reaching up to 9.1\% and 6.1\% improvements on math and coding tasks respectively, and performs on par with the LoRA finetuned model and ReasonFlux-1.5B PRM. 

\begin{table}[H]
\scriptsize
\setlength{\tabcolsep}{3pt}
\renewcommand{\arraystretch}{1.1}
\centering
\resizebox{\textwidth}{!}{

\begin{tabular}{ccc|c|cccccc|cc}
\toprule
Model & Calibration set & Test & \method  
& \makecell{Best \\ Confidence-based\\ Strategy} 
& LoRA  
& PRM 
& \method-WMV  
& Oracle 
& Delta & Pct. gap closed \\
\midrule
GPT-OSS-20B & MATH-500 & AIME & 89.8 $\pm$ 2.9 
& 93.3 $\pm$ 0.0 & \textbf{94.8 $\pm$ 2.4} & 92.2 $\pm$ 1.9 & 94.0 $\pm$ 1.3 & 97.8 
& \cellcolor{red!25}-3.5 & -77.8 \\
GPT-OSS-20B & AIME & MATH-500 & 95.0 $\pm$ 0.5
& 94.6 $\pm$ 0.9 & 94.8 $\pm$ 0.3 & 95.5 $\pm$ 0.3 & \textbf{96.1 $\pm$ 0.3} & 97.7 
& \cellcolor{green!15}+0.4 & 12.9 \\
Qwen-1.7B & MATH-500 & AIME & 48.9 $\pm$ 3.1
& 52.2 $\pm$ 5.1 & 50.4 $\pm$ 5.9 & 41.1 $\pm$ 8.4 & 55.2 $\pm$ 2.9 & 68.9
& \cellcolor{red!25}-3.3 & -19.8 \\
Qwen-1.7B & AIME & MATH-500 & 88.3 $\pm$ 0.7
& 88.1 $\pm$ 0.4 & 88.9 $\pm$ 0.5 & 87.3 $\pm$ 0.8 & 90.6 $\pm$ 0.4 & 94.9
& \cellcolor{green!15}+0.2 & 2.9 \\
Qwen-30B-A3B & MATH-500 & AIME & \textbf{42.4 $\pm$ 5.5} 
& 33.3 $\pm$ 3.3 & — & 37.8 $\pm$ 5.1 & 45.7 $\pm$ 3.8 & 57.8 
& \cellcolor{green!45}+9.1 & 37.1 \\
Qwen-30B-A3B & AIME & MATH-500 & \textbf{91.3 $\pm$ 0.5} 
& 85.0 $\pm$ 1.6 & — & 88.9 $\pm$ 0.2 & 91.6 $\pm$ 0.3 & 93.2 
& \cellcolor{green!25}+6.3 & 76.8 \\
\bottomrule
\end{tabular}
}
\vspace{1em}
\caption{Best-of-$N$ results for math domain. All values are mean $\pm$ standard deviation. \method{} consistently matches or outperforms strong baselines and closes a substantial fraction of the gap to oracle selection, while remaining computationally efficient.}
\vspace{-2mm}
\label{tab:main_results_math}
\end{table}

\begin{table}[H]
\scriptsize
\setlength{\tabcolsep}{3pt}
\renewcommand{\arraystretch}{1.1}
\centering
\resizebox{\textwidth}{!}{
\begin{tabular}{ccc|c|ccccc|cc}
\toprule
Model & Calibration set & Test & \method 
& \makecell{Best \\ Confidence-based\\ Strategy} 
& LoRA  
& PRM 
& Oracle 
& Delta & Pct. gap closed (\%) \\
\midrule
GPT-OSS-20B & KodCode & HumanEval & \textbf{93.2 $\pm$ 0.6} 
& 91.7 $\pm$ 0.9 & 92.3 $\pm$ 0.4 & 89.8 $\pm$ 2.5 & 93.9 
& \cellcolor{green!25}+1.5 & 68.2 \\
GPT-OSS-20B & HumanEval & KodCode & \textbf{86.8 $\pm$ 0.7} 
& 83.8 $\pm$ 0.9 & 85.2 $\pm$ 1.0 & 82.5 $\pm$ 4.0 & 94.5 
& \cellcolor{green!25}+3.0 & 28.0 \\
OLMo-2-7B & KodCode & HumanEval & 35.0 $\pm$ 1.7
& 31.7 $\pm$ 1.1 & \textbf{37.7 $\pm$ 2.7} & 32.9 $\pm$ 1.6 & 73.4 
& \cellcolor{green!25}+3.3 & 7.9 \\
OLMo-2-7B & HumanEval & KodCode & \textbf{33.1 $\pm$ 0.9} 
& 27.0 $\pm$ 0.2 & 32.7 $\pm$ 0.8 & 29.3 $\pm$ 1.3 & 66.7 
& \cellcolor{green!35}+6.1 & 15.4 \\
Qwen-1.7B & KodCode & HumanEval & 59.3 $\pm$ 8.1 
& 56.1 $\pm$ 2.1 & \textbf{63.8 $\pm$ 9.2} & 60.4 $\pm$ 4.3 & 87.4 
& \cellcolor{green!25}+3.2 & 10.2 \\
Qwen-1.7B & HumanEval & KodCode & 67.1 $\pm$ 12.2 
& 61.3 $\pm$ 11.6 & \textbf{68.3 $\pm$ 9.1} & 64.5 $\pm$ 6.9 & 90.8 
& \cellcolor{green!35}+5.8 & 19.7 \\
Qwen-30B-A3B & KodCode & HumanEval & \textbf{90.4 $\pm$ 0.8} 
& 90.0 $\pm$ 0.7 & — & 87.6 $\pm$ 0.7 & 92.7 
& \cellcolor{green!15}+0.4 & 14.8 \\
Qwen-30B-A3B & HumanEval & KodCode & \textbf{88.2 $\pm$ 0.6} 
& 87.0 $\pm$ 1.0 & —  & 84.6 $\pm$ 1.1 & 93.7 
& \cellcolor{green!15}+1.2 & 17.9 \\
\bottomrule
\end{tabular}
}
\vspace{1em}
\caption{Best-of-$N$ results for coding domain. All values are mean $\pm$ standard deviation. \method{} consistently matches or outperforms strong baselines and closes a substantial fraction of the gap to oracle selection, while remaining computationally efficient.}
\vspace{-2mm}
\label{tab:main_results_code}
\end{table}

We further compare the efficiency of \method~against LoRA in \autoref{tab:scatr_efficiency_comparison_appendix} and \autoref{tab:scatr_efficiency_comparison_lora_t2}. 
Despite its significantly smaller footprint and training and inference times, \method~performs on par with significantly larger models, demonstrating a markedly better efficiency–performance trade-off.

Finally, we compare the efficiency and accuracy of ReasonFlux-PRM-1.5B~\citep{zou2025reasonfluxprmtrajectoryawareprmslong} and \method{} across evaluation datasets for mathematical reasoning tasks. \method{} achieves comparable or higher accuracy while using orders of magnitude fewer parameters and providing substantial inference speedups of up to $1000\times$. Results are shown in \autoref{tab:scatr_efficiency_comparison_math_reasonflux}.

\begin{table}[H]
\centering
\scriptsize
\begin{tabular}{lllrrrrrr}
\toprule
 &  &   \multicolumn{3}{c}{LoRA} & \multicolumn{3}{c}{SCATR (ours)} \\
\cmidrule(lr){3-5} \cmidrule(lr){6-8}
Model & Dataset & TFLOPs & Train (min) & Acc. & TFLOPs & Train (min) & Acc. \\
\midrule

\multirow{4}{*}{GPT-OSS}
& AIME      & $12690\times$ & $119.07$ & $0.95$ & $8.7\times10^{1}$ & $2.17$ & $0.90$ \\
& HumanEval  & $13268\times$ & $206.29$ & $0.92$ & $1.5\times10^{2}$ & $5.56$ & $0.93$ \\
& KodCode    & $15310\times$ & $35.78$ & $0.85$ & $2.4\times10^{1}$ & $1.61$ & $0.87$ \\
& MATH-500   & $108503\times$ & $74.37$ & $0.95$ & $7.4\times10^{0}$ & $0.47$ & $0.95$ \\
\midrule
\multirow{2}{*}{OLMo-2-7B}
 & HumanEval    & $2075\times$ & $35.38$ & $0.38$ & $1.6\times10^{2}$ & $3.89$ & $0.35$ \\
 & KodCode      & $2515\times$ & $5.76$ & $0.33$ & $1.9\times10^{1}$ & $1.37$ & $0.33$ \\
 \midrule
\multirow{4}{*}{Qwen-1.7B }
& AIME        & $2324\times$ & $29.50$ & $0.50$ & $8.9\times10^{1}$ & $2.24$ & $0.49$ \\
& HumanEval    & $5374\times$ & $49.32$ & $0.64$ & $7.4\times10^{1}$ & $6.48$ & $0.59$ \\
& KodCode      & $3217\times$ & $7.90$ & $0.68$ & $2.0\times10^{1}$ & $1.52$ & $0.67$ \\
& MATH-500     & $39568\times$ & $30.02$ & $0.89$ & $5.4\times10^{0}$ & $0.49$ & $0.88$ \\
\bottomrule
\end{tabular}
\vspace{1em}
\caption{Comparison of LoRA finetuned model and SCATR (ours) across different models and datasets. 
The table reports training size (TFLOPs), training time in minutes, and accuracy for each method. SCATR demonstrates competitive performance with significantly reduced computational requirements.}
\label{tab:scatr_efficiency_comparison_appendix}
\end{table}

\begin{table}[H]
\centering
\scriptsize
\setlength{\tabcolsep}{3pt}
\renewcommand{\arraystretch}{0.8}
\vspace{-2mm}
\begin{tabular}{ll|ccc|ccccc}
\toprule
\multirow{2}{*}{\textbf{Model}} & \multirow{2}{*}{\textbf{Dataset}} & \multicolumn{3}{c|}{\textbf{LoRA}} & \multicolumn{5}{c}{\textbf{\method~ (ours)}} \\
\cmidrule(lr){3-5} \cmidrule(lr){6-10}
 & & \textbf{Size} & \textbf{Infer (ms) $\downarrow$} & \textbf{Acc. $\uparrow$} & \textbf{Size} & \textbf{Infer (ms) $\downarrow$} & \textbf{Speedup $\uparrow$} & \textbf{Acc. $\uparrow$} & \textbf{$\Delta$ Accuracy $\uparrow$} \\
\midrule
\multirow{4}{*}{GPT-OSS-20B}
& AIME      & 20B & 159.9 & 94.8$\pm$2.4 & 6.7M & 0.3 & \textcolor{darkgreen2}{625.5$\times$}  & 89.8$\pm$2.9 & \cellcolor{red!10}{-5.1} \\
& HumanEval & 20B & 249.5 & 92.3$\pm$0.4 & 5.0M & 0.2 & \textcolor{darkgreen2}{1352.7$\times$} & 93.2$\pm$0.6 & \cellcolor{green!10}{+0.9} \\
& KodCode   & 20B & 257.6 & 85.2$\pm$1.0 & 2.9M & 0.2 & \textcolor{darkgreen2}{1271.9$\times$} & 86.8$\pm$0.7 & \cellcolor{green!10}{+1.6} \\
& MATH-500  & 20B & 616.1 & 94.8$\pm$0.3 & 2.3M & 0.2 & \textcolor{darkgreen2}{3534.1$\times$} & 95.0$\pm$0.5 & \cellcolor{green!5}{+0.2} \\
\midrule
\multirow{2}{*}{OLMo-2-7B}

& HumanEval & 7B  & 48.5  & 37.7$\pm$2.7 & 6.6M & 0.2 & \textcolor{darkgreen2}{305.5$\times$} & 35.0$\pm$1.7 & \cellcolor{red!10}{-2.7} \\
& KodCode   & 7B  & 49.2  & 32.7$\pm$0.8 & 2.2M & 0.2 & \textcolor{darkgreen2}{251.0$\times$} & 33.1$\pm$0.9 & \cellcolor{green!5}{+0.4} \\

\midrule
\multirow{4}{*}{Qwen-1.7B}
& AIME      & 1.7B  & 55.3  & 50.4$\pm$5.9  & 6.3M & 0.2 & \textcolor{darkgreen2}{326.0$\times$} & 48.9$\pm$3.1  & \cellcolor{red!10}{-1.5} \\
& HumanEval & 1.7B  & 47.0  & 63.8$\pm$9.2  & 2.6M & 0.2 & \textcolor{darkgreen2}{305.2$\times$} & 59.3$\pm$8.1  & \cellcolor{red!10}{-4.5} \\
& KodCode   & 1.7B  & 48.7  & 68.3$\pm$9.1  & 2.2M & 0.2 & \textcolor{darkgreen2}{270.5$\times$} & 67.1$\pm$12.2 & \cellcolor{red!5}{-1.2} \\
& MATH-500  & 1.7B  & 66.9  & 88.9$\pm$0.5  & 1.7M & 0.2 & \textcolor{darkgreen2}{436.1$\times$} & 88.3$\pm$0.7  & \cellcolor{red!5}{-0.6} \\

\bottomrule
\end{tabular}
\vspace{1em}
\caption{Efficiency and accuracy comparison of LoRA and \method~ across evaluation datasets. \method~ achieves comparable accuracy while using orders of magnitude fewer parameters and providing substantial inference speedups.}
\label{tab:scatr_efficiency_comparison_lora_t2}
\end{table}

\begin{table}[H]
\centering
\scriptsize
\setlength{\tabcolsep}{3pt}
\renewcommand{\arraystretch}{1.1}
\begin{tabular}{ll|ccc|ccccc}
\toprule
\multirow{2}{*}{Model} & \multirow{2}{*}{Dataset} & \multicolumn{3}{c|}{ReasonFlux-PRM-1.5B} & \multicolumn{5}{c}{\method~ (ours)} \\
\cmidrule(lr){3-5} \cmidrule(lr){6-10}
 & & Size & Infer (ms) & Acc. & Size & Infer (ms) & Speedup & Acc. & $\Delta$Acc. \\
\midrule

\multirow{2}{*}{GPT-OSS-20B}
& AIME  & 1.5B & 175.8 & 92.2$\pm$1.9 & 6.7M & 0.26 & \textcolor{darkgreen}{687.5$\times$} & 89.8$\pm$2.9 & \textcolor{red}{-2.5} \\
& MATH-500  & 1.5B & 119.3 & 95.5$\pm$0.3 & 2.3M & 0.17 & \textcolor{darkgreen}{684.3$\times$} & 95.0$\pm$0.5 & \textcolor{red}{-0.5} \\

\midrule
\multirow{2}{*}{Qwen-1.7B}
& AIME & 1.5B & 188.7 & 41.1$\pm$8.4 & 6.3M & 0.17 & \textcolor{darkgreen}{1113.3$\times$} & 48.9$\pm$3.1 & \textcolor{darkgreen}{+7.8} \\
& MATH-500 & 1.5B & 170.9 & 87.3$\pm$0.8 & 1.7M & 0.15 & \textcolor{darkgreen}{1115.2$\times$} & 88.3$\pm$0.7 & \textcolor{darkgreen}{+1.0} \\

\midrule
\multirow{2}{*}{Qwen-30B-A3B}
& AIME & 1.5B & 175.0 & 37.8$\pm$5.1 & 2.7M & 0.3 & \textcolor{darkgreen}{691.3$\times$} & 42.4$\pm$5.5 & \textcolor{darkgreen}{+4.6} \\
& MATH-500 & 1.5B & 167.8 & 88.9$\pm$0.2 & 2.6M & 0.1 & \textcolor{darkgreen}{1148.1$\times$} & 91.3$\pm$0.5 & \textcolor{darkgreen}{+2.3} \\
\bottomrule
\end{tabular}
\vspace{1em}
\caption{Efficiency and accuracy comparison of ReasonFlux-PRM-1.5B~\citep{zou2025reasonfluxprmtrajectoryawareprmslong} and \method~ across evaluation datasets for math tasks. SCATR achieves comparable or higher accuracy while using orders of magnitude fewer parameters and providing substantial inference speedups of up to 1000$\times$.}
\label{tab:scatr_efficiency_comparison_math_reasonflux}
\end{table}

\begin{figure}[]
    \centering
    \includegraphics[width=\linewidth]{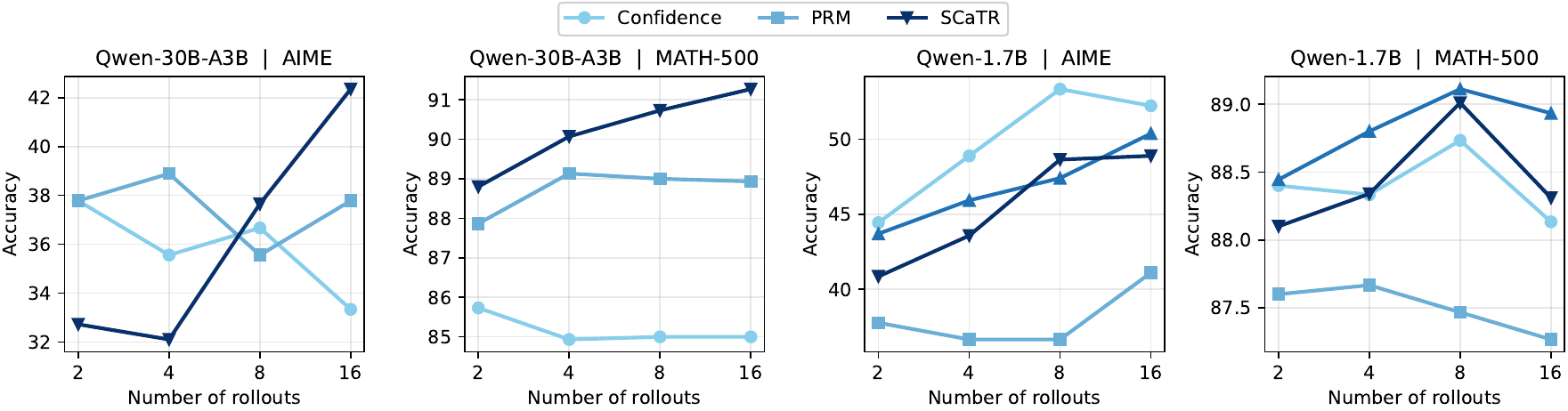} 
    \caption{Accuracy as a function of the number of rollouts across two models and two math datasets. For each experiment, the calibration set is drawn from the alternate dataset. We compare four selection strategies: Confidence, the strongest confidence-based baseline selecting the rollout with the highest confidence score; PRM, using a fixed ReasonFlux-1.5B process reward model; LoRA, a fine-tuned model trained to predict rollout correctness.}
    \label{fig:performance_num_rollouts_math}
\end{figure}

\subsection{Confidence-Based Methods: Additional Results}
\label{sec:metric_experiments_results}

\begin{figure}[]
    \centering
    \includegraphics[width=0.95\linewidth]{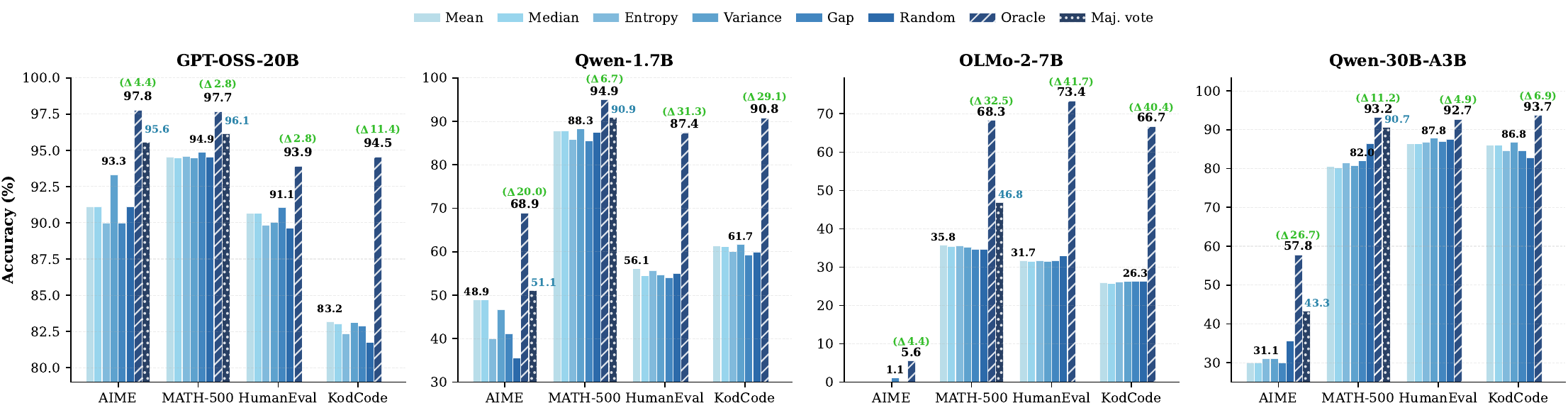}
    \caption{Performance of tail-aggregated token-level uncertainty metrics compared with random, majority vote and oracle best-of-N responses selection. Metrics include standard mean confidence ($C_i$), median and variance of top-$k$ log-probabilities, probability gap between the first two top-$k$ probabilities, and Shannon entropy, computed from the normalized distribution of top-$k$ log-probabilities at each token position and aggregated over the tail of each sequence. Confidence-based metrics, including the standard mean ($C_i$), behave similarly to random selection and remain far below oracle performance, defined as success whenever any candidate response is correct, indicating that fixed confidence measures provide a limited signal for effective response selection.\looseness=-1}
    \label{fig:metrics_gpt_qwen30}
\end{figure}

We compare tail-aggregated token-level uncertainty metrics on GPT-OSS-20B and Qwen-30B, as shown in~\autoref{fig:metrics_gpt_qwen30}, complementing the results in the main paper that cover additional model families. Consistent with prior findings, all confidence- and uncertainty-based metrics perform close to random selection and remain substantially below oracle performance, with no metric exhibiting reliable advantages across datasets. These results reinforce the conclusion that token-level uncertainty signals provide limited guidance for effective Best-of-$N$ response selection.

\begin{figure}[h]
    \centering
    \includegraphics[width=0.95\linewidth]{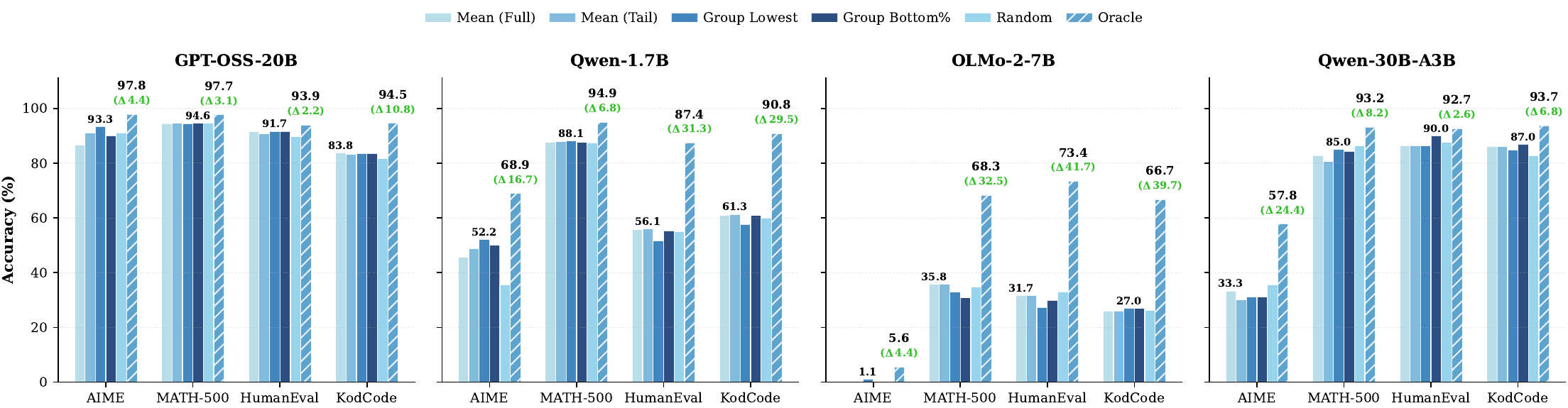}
    \caption{Performance of confidence-based Best-of-$N$ selection using different aggregation strategies for token-level confidence, compared with random and oracle selection. Aggregation methods include confidence averaged over the full sequence (Full), over the final 2048 tokens (Tail), selection based on the highest lowest group confidence with group size 1024 (Group Lowest), and selection based on the highest bottom 10\% group confidence (Bottom 10\%). Across models and datasets, different aggregation strategies yield similar performance and remain far below oracle performance—defined as success whenever any candidate response is correct—exhibiting only minor variations and no consistent pattern, indicating that no single aggregation strategy consistently yields optimal responses, leaving it unclear which confidence-based criterion should be preferred in practice.}
    \label{fig:aggregation_strategies}
\end{figure}

We evaluate confidence-based Best-of-$N$ selection by applying multiple aggregation strategies to token-level confidence scores, including full-sequence averaging, tail-only aggregation, and group-based criteria that emphasize low-confidence regions of the trace. As shown in ~\autoref{fig:aggregation_strategies}, all aggregation strategies yield similar performance across models and datasets, with results close to random selection and substantially below oracle performance. While minor variations are observed depending on the model and dataset, no aggregation strategy consistently outperforms the others. These findings suggest that fixed confidence-based aggregation provides limited signal for identifying correct reasoning traces and offers no clear guidance on which aggregation strategy should be preferred in practice.

\begin{figure}[]
    \centering
    \includegraphics[width=0.8\linewidth]{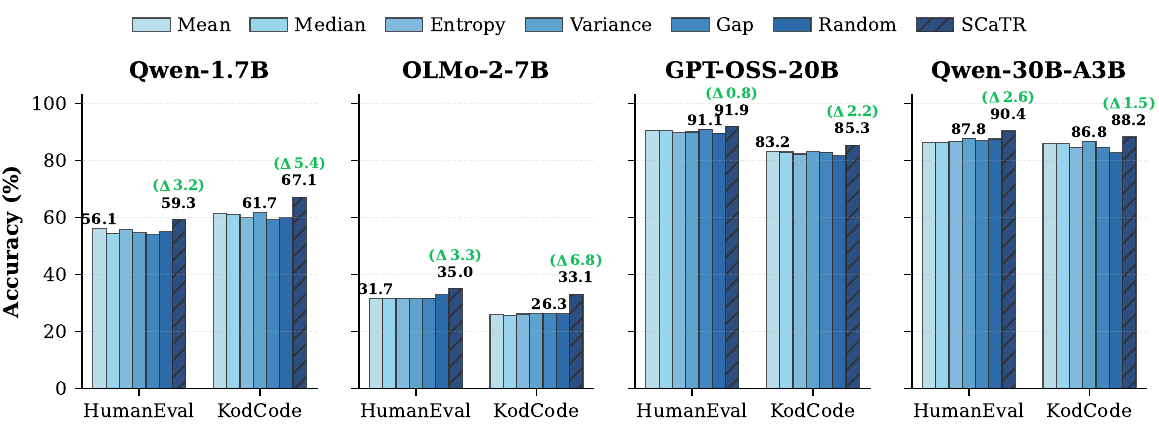}
    \caption{Performance of tail-aggregated token-level uncertainty metrics compared to random best-of-$N$ response selection and \method~ across all models in the coding domain. Metrics include standard mean confidence ($C_i$)~\citep{fu2025deepthinkconfidence}, median and variance of top-$k$ log-probabilities, probability gap between the top two tokens, and Shannon entropy, computed from normalized top-$k$ distributions and aggregated over the sequence tail (see ~\autoref{app:confidence_definitions_and_metrics}). \method~consistently outperforms all confidence-based metrics, which perform close to random selection, highlighting the limitations of fixed uncertainty measures. }
    \label{fig:scatr_vs_uncertainty_metrics_coding_datasets}
\end{figure}

Finally, a comprehensive overview of tail-aggregated token-level uncertainty metrics compared to random best-of-$N$ response selection and \method{} across all models in the coding domain (extending the results shown in the main paper) is presented in ~\autoref{fig:scatr_vs_uncertainty_metrics_coding_datasets}. \method{} consistently outperforms all confidence-based metrics, which perform close to random selection, highlighting the limitations of fixed uncertainty metrics.

\subsection{Ablation Studies}
\label{app:extended_ablations_section}
\subsubsection{Layer choice analysis}\label{sec:layer}
Here, we experiment with using the hidden states from different intermediate layers, rather than just the last (which \method~ uses). We consider layers at constant proportions of the total number of layers (taking floors if the resulting value is non-integral). We see that performance is similar across layers, with the last commonly exhibiting lower variance. Below are Best-of-$N$ results, shown for $N=4, 8, 16$ 
in \autoref{tab:layer_n4}, \autoref{tab:layer_n8} and \autoref{tab:layer_analysis_main}, 
respectively.

\begin{table}[h]
\centering
\resizebox{\linewidth}{!}{%
{
\footnotesize
\setlength{\tabcolsep}{4pt}
\begin{tabular}{lllccccccc}
\toprule
\textbf{Model} & \textbf{Train} & \textbf{Test} & \textbf{$.15L$} & \textbf{$.3L$} & \textbf{$.45L$} & \textbf{$.6L$} & \textbf{$.75L$} & \textbf{$.9L$} & \textbf{$L-1$} \\
\midrule
GPT-OSS-20B & HumanEval & KodCode & 85.0 $\pm$ 2.8 & 85.0 $\pm$ 2.7 & 85.2 $\pm$ 2.4 & 85.0 $\pm$ 2.4 & 86.0 $\pm$ 1.5 & \textbf{86.3 $\pm$ 1.4} & 85.9 $\pm$ 1.1 \\
GPT-OSS-20B & KodCode & HumanEval & \textbf{92.7 $\pm$ 0.6} & 92.6 $\pm$ 0.6 & 92.5 $\pm$ 0.8 & 92.5 $\pm$ 0.7 & 92.5 $\pm$ 0.7 & 92.3 $\pm$ 1.3 & 92.6 $\pm$ 1.0 \\
\midrule
Qwen-30B-A3B & HumanEval & KodCode & 87.0 $\pm$ 0.6 & 86.7 $\pm$ 0.6 & 86.9 $\pm$ 0.7 & 86.9 $\pm$ 0.6 & 86.7 $\pm$ 0.6 & 86.8 $\pm$ 0.7 & \textbf{87.1 $\pm$ 0.5} \\
Qwen-30B-A3B & KodCode & HumanEval & 89.4 $\pm$ 1.0 & \textbf{89.7 $\pm$ 0.9} & 89.6 $\pm$ 0.9 & 89.5 $\pm$ 1.0 & 89.7 $\pm$ 1.1 & 89.3 $\pm$ 1.1 & 89.3 $\pm$ 1.2 \\
\midrule
Qwen-1.7B & HumanEval & KodCode & 60.8 $\pm$ 14.7 & 61.1 $\pm$ 14.7 & 60.6 $\pm$ 15.3 & \textbf{61.2 $\pm$ 14.6} & 60.8 $\pm$ 15.0 & 60.8 $\pm$ 14.9 & 61.0 $\pm$ 14.7 \\
Qwen-1.7B & KodCode & HumanEval & 55.0 $\pm$ 6.1 & \textbf{55.2 $\pm$ 6.2} & 55.1 $\pm$ 6.2 & 55.0 $\pm$ 6.5 & 54.9 $\pm$ 6.6 & 54.4 $\pm$ 6.6 & 54.5 $\pm$ 7.0 \\
\midrule
OLMo-2-7B & HumanEval & KodCode & 29.7 $\pm$ 0.6 & 31.1 $\pm$ 0.9 & 30.6 $\pm$ 1.1 & \textbf{31.5 $\pm$ 1.1} & 31.1 $\pm$ 0.8 & 31.2 $\pm$ 0.9 & 31.3 $\pm$ 0.8 \\
OLMo-2-7B & KodCode & HumanEval & 32.7 $\pm$ 1.6 & 32.5 $\pm$ 1.3 & 32.9 $\pm$ 2.7 & 34.3 $\pm$ 1.7 & 33.4 $\pm$ 1.5 & \textbf{34.4 $\pm$ 1.9} & 33.7 $\pm$ 2.1 \\
\bottomrule
\end{tabular}
}
}
\vspace{1em}
\caption{Comparison of Best-of-$N$ accuracies ($N=4$) when using different intermediate layers for \method~ (where we floor the respective percentages).}

\label{tab:layer_n4}
\end{table}

\begin{table}[h]
\centering
\resizebox{\linewidth}{!}{%
{
\footnotesize
\setlength{\tabcolsep}{4pt}
\begin{tabular}{lllccccccc}
\toprule
\textbf{Model} & \textbf{Train} & \textbf{Test} & \textbf{$.15L$} & \textbf{$.3L$} & \textbf{$.45L$} & \textbf{$.6L$} & \textbf{$.75L$} & \textbf{$.9L$} & \textbf{$L-1$} \\
\midrule
GPT-OSS-20B & HumanEval & KodCode & 84.7 $\pm$ 3.2 & 84.7 $\pm$ 3.1 & 85.4 $\pm$ 2.6 & 85.0 $\pm$ 2.5 & 86.4 $\pm$ 1.5 & \textbf{86.8 $\pm$ 1.2} & 86.6 $\pm$ 0.6 \\
GPT-OSS-20B & KodCode & HumanEval & \textbf{92.7 $\pm$ 0.6} & 92.6 $\pm$ 0.8 & 92.3 $\pm$ 0.9 & 92.5 $\pm$ 0.9 & 92.4 $\pm$ 0.9 & 92.1 $\pm$ 0.8 & 92.7 $\pm$ 0.8 \\
\midrule
Qwen-30B-A3B & HumanEval & KodCode & 87.1 $\pm$ 0.7 & 86.9 $\pm$ 0.7 & 87.1 $\pm$ 1.0 & 86.9 $\pm$ 0.7 & 87.3 $\pm$ 0.7 & 87.0 $\pm$ 0.8 & \textbf{87.7 $\pm$ 0.5} \\
Qwen-30B-A3B & KodCode & HumanEval & 90.5 $\pm$ 0.7 & \textbf{91.3 $\pm$ 0.8} & 90.5 $\pm$ 0.9 & 90.5 $\pm$ 0.8 & 90.4 $\pm$ 0.8 & 90.2 $\pm$ 0.7 & 90.2 $\pm$ 0.8 \\
\midrule
Qwen-1.7B & HumanEval & KodCode & 61.5 $\pm$ 15.4 & 61.8 $\pm$ 15.7 & 61.2 $\pm$ 16.2 & 62.3 $\pm$ 15.5 & 61.7 $\pm$ 15.7 & 61.6 $\pm$ 15.6 & \textbf{62.4 $\pm$ 14.8} \\
Qwen-1.7B & KodCode & HumanEval & 54.2 $\pm$ 9.7 & 54.1 $\pm$ 9.2 & \textbf{54.3 $\pm$ 9.5} & 53.8 $\pm$ 9.5 & 53.1 $\pm$ 9.1 & 52.6 $\pm$ 8.4 & 52.6 $\pm$ 9.4 \\
\midrule
OLMo-2-7B & HumanEval & KodCode & 30.2 $\pm$ 1.2 & 31.6 $\pm$ 1.0 & 31.6 $\pm$ 1.1 & 32.4 $\pm$ 1.4 & 32.0 $\pm$ 1.1 & \textbf{32.4 $\pm$ 1.2} & 32.2 $\pm$ 0.9 \\
OLMo-2-7B & KodCode & HumanEval & 31.7 $\pm$ 1.8 & 34.4 $\pm$ 1.7 & 34.1 $\pm$ 2.4 & \textbf{35.8 $\pm$ 1.5} & 34.5 $\pm$ 1.5 & 35.6 $\pm$ 1.4 & 35.2 $\pm$ 2.0 \\
\bottomrule
\end{tabular}
}
}
\vspace{1em}
\caption{Comparison of Best-of-$N$ accuracies ($N=8$) when using different intermediate layers for \method.}

\label{tab:layer_n8}
\end{table}

\begin{table}[h]
\centering

\resizebox{\linewidth}{!}{%
{
\small
\setlength{\tabcolsep}{4pt}
\begin{tabular}{lllccccccc}
\toprule
\textbf{Model} & \textbf{Train} & \textbf{Test} & \textbf{$.15L$} & \textbf{$.3L$} & \textbf{$.45L$} & \textbf{$.6L$} & \textbf{$.75L$} & \textbf{$.9L$} & \textbf{$L-1$} \\
\midrule
GPT-OSS-20B & HumanEval & KodCode & 84.6 $\pm$ 4.3 & 84.4 $\pm$ 4.3 & 85.0 $\pm$ 3.4 & 84.7 $\pm$ 3.2 & 86.4 $\pm$ 1.7 & \textbf{86.8 $\pm$ 1.2} & 86.8 $\pm$ 0.7 \\
GPT-OSS-20B & KodCode & HumanEval & 93.0 $\pm$ 0.6 & 93.2 $\pm$ 0.6 & \textbf{93.2 $\pm$ 0.4} & 93.1 $\pm$ 0.7 & 93.1 $\pm$ 0.5 & 92.9 $\pm$ 0.7 & 93.2 $\pm$ 0.6 \\
\midrule
Qwen-30B-A3B & HumanEval & KodCode & 87.5 $\pm$ 0.8 & 87.1 $\pm$ 1.0 & 87.4 $\pm$ 1.3 & 87.2 $\pm$ 1.0 & 87.5 $\pm$ 0.8 & 87.4 $\pm$ 1.0 & \textbf{88.3 $\pm$ 0.6} \\
Qwen-30B-A3B & KodCode & HumanEval & 90.4 $\pm$ 0.7 & 91.0 $\pm$ 0.7 & 90.4 $\pm$ 0.7 & \textbf{91.1 $\pm$ 0.7} & 90.8 $\pm$ 0.7 & 90.5 $\pm$ 0.8 & 90.4 $\pm$ 0.8 \\
\midrule
Qwen-1.7B & HumanEval & KodCode & 65.2 $\pm$ 13.9 & 66.2 $\pm$ 13.6 & 65.6 $\pm$ 14.5 & \textbf{67.4 $\pm$ 12.8} & 66.2 $\pm$ 13.6 & 66.4 $\pm$ 13.2 & 67.1 $\pm$ 12.2 \\
Qwen-1.7B & KodCode & HumanEval & 60.6 $\pm$ 8.1 & 60.6 $\pm$ 7.3 & \textbf{61.7 $\pm$ 7.6} & 59.6 $\pm$ 7.0 & 59.6 $\pm$ 7.8 & 60.3 $\pm$ 7.3 & 59.3 $\pm$ 8.1 \\
\midrule
OLMo-2-7B & HumanEval & KodCode & 30.0 $\pm$ 0.7 & 32.3 $\pm$ 0.9 & 32.2 $\pm$ 1.3 & 33.0 $\pm$ 1.1 & 32.8 $\pm$ 0.9 & \textbf{33.2 $\pm$ 1.2} & 33.1 $\pm$ 0.9 \\
OLMo-2-7B & KodCode & HumanEval & 30.6 $\pm$ 1.6 & 33.0 $\pm$ 2.0 & 34.3 $\pm$ 3.0 & \textbf{36.5 $\pm$ 2.7} & 34.2 $\pm$ 2.3 & 35.5 $\pm$ 1.6 & 35.0 $\pm$ 1.7 \\
\bottomrule
\end{tabular}
}
}
\vspace{1em}
\caption{Comparison of Best-of-$N$ accuracies ($N=16$) when using different intermediate layers for \method.}
\label{tab:layer_analysis_main}
\end{table}

\subsubsection{Calibration Set Size analysis}\label{sec:fraction}
Here, we experiment with using just a fraction of the calibration set to train the scoring function. We see that additional data could be helpful for \method~ trained on HumanEval (164 problems), but for Kodcode (1000 problems, which has been subsetted), performance plateaus. Below are results, evaluating \method~ on $N=4, 8, 12, 16$ rollouts as shown in \autoref{tab:frac_main}.
\begin{table}[h]
\centering

\resizebox{0.8\linewidth}{!}{%
{
\small
\setlength{\tabcolsep}{4pt}
\begin{tabular}{lllcccc}
\toprule
\textbf{Model} & \textbf{Train} & \textbf{Test} & \textbf{$10\%$} & \textbf{$25\%$} & \textbf{$50\%$} & \textbf{$100\%$} \\
\midrule
GPT-OSS-20B & HumanEval & KodCode & 78.8 $\pm$ 6.6 & 84.7 $\pm$ 2.8 & 85.6 $\pm$ 2.4 & 86.8 $\pm$ 0.7 \\
GPT-OSS-20B & KodCode & HumanEval & 92.4 $\pm$ 0.7 & 92.6 $\pm$ 0.9 & 92.9 $\pm$ 0.8 & 93.2 $\pm$ 0.6 \\
\midrule
Qwen-30B-A3B & HumanEval & KodCode & 85.5 $\pm$ 4.6 & 87.7 $\pm$ 0.6 & 87.8 $\pm$ 0.4 & 88.3 $\pm$ 0.6 \\
Qwen-30B-A3B & KodCode & HumanEval & 89.7 $\pm$ 2.0 & 90.0 $\pm$ 1.3 & 90.3 $\pm$ 1.0 & 90.4 $\pm$ 0.8 \\
\midrule
Qwen-1.7B & HumanEval & KodCode & 64.2 $\pm$ 13.4 & 65.9 $\pm$ 10.9 & 66.2 $\pm$ 13.5 & 67.1 $\pm$ 12.2 \\
Qwen-1.7B & KodCode & HumanEval & 59.1 $\pm$ 7.5 & 59.7 $\pm$ 8.0 & 59.4 $\pm$ 7.8 & 59.3 $\pm$ 8.1 \\
\midrule
OLMo-2-7B & HumanEval & KodCode & 29.5 $\pm$ 1.6 & 30.3 $\pm$ 2.6 & 32.7 $\pm$ 1.0 & 33.1 $\pm$ 0.9 \\
OLMo-2-7B & KodCode & HumanEval & 33.7 $\pm$ 2.2 & 34.9 $\pm$ 1.6 & 35.2 $\pm$ 2.3 & 35.0 $\pm$ 1.7 \\
\bottomrule
\end{tabular}
}
}
\vspace{1em}
\caption{\method~Best-of-$N$ results with $N=16$. Here, we train on a fraction of the calibration set: (10\%, 25\%, 50\%, 100\%). We see that \method~ trained on HumanEval may benefit from more data, while training on KodCode seems to plateau by the time we use the full dataset.}

\label{tab:frac_main}
\end{table}

\subsubsection{Scoring Function Design analysis}\label{sec:transformer}

Here, we present the full results for the scoring function design ablation, where we try more sophisticated scoring function designs, some transformer based. We show results here for additional models, evaluated on $N=4, 8, 12$ in \autoref{tab:ablation_n4}, \autoref{tab:ablation_n8}, and 
\autoref{tab:ablation_n12}, respectively.

\begin{table}[H]
\centering

\resizebox{\linewidth}{!}{%
{
\footnotesize
\setlength{\tabcolsep}{4pt}
\begin{tabular}{lll|ccccc|c}
\toprule
\textbf{Model} & \textbf{Train} & \textbf{Test} & \textbf{Final} & \textbf{Last-10} & \textbf{Special} & \textbf{Ensemble} & \textbf{Best Layer} & \textbf{\method} \\
\midrule
Qwen-30B-A3B & HumanEval & KodCode & \textbf{87.4 $\pm$ 0.3} & 86.8 $\pm$ 0.6 & 87.2 $\pm$ 0.5 & 86.9 $\pm$ 0.5 & 86.7 $\pm$ 0.7 & 87.1 $\pm$ 0.5 \\
Qwen-30B-A3B & KodCode & HumanEval & 89.3 $\pm$ 1.4 & \textbf{89.8 $\pm$ 1.0} & 89.4 $\pm$ 1.1 & 89.6 $\pm$ 1.0 & 89.6 $\pm$ 1.1 & 89.3 $\pm$ 1.2 \\
\midrule
Qwen-1.7B & HumanEval & KodCode & \textbf{61.8 $\pm$ 14.6} & 61.2 $\pm$ 13.9 & 57.7 $\pm$ 14.1 & 61.0 $\pm$ 15.1 & 61.2 $\pm$ 15.0 & 61.0 $\pm$ 14.7 \\
Qwen-1.7B & KodCode & HumanEval & \textbf{55.5 $\pm$ 5.8} & 54.4 $\pm$ 6.0 & 53.6 $\pm$ 3.5 & 55.2 $\pm$ 6.3 & 55.2 $\pm$ 6.3 & 54.5 $\pm$ 7.0 \\
\midrule
GPT-OSS-20B & HumanEval & KodCode & 85.8 $\pm$ 1.3 & 85.3 $\pm$ 1.1 & 85.5 $\pm$ 1.7 & \textbf{86.0 $\pm$ 2.1} & 85.5 $\pm$ 2.0 & 86.0 $\pm$ 1.3 \\
GPT-OSS-20B & KodCode & HumanEval & 92.5 $\pm$ 0.9 & 92.5 $\pm$ 1.0 & 92.6 $\pm$ 0.9 & \textbf{92.8 $\pm$ 0.7} & 92.8 $\pm$ 0.6 & 92.5 $\pm$ 0.9 \\
\midrule
OLMo-2-7B & HumanEval & KodCode & 31.6 $\pm$ 0.8 & 29.0 $\pm$ 0.9 & 28.9 $\pm$ 0.8 & \textbf{32.0 $\pm$ 0.9} & 31.2 $\pm$ 1.0 & 31.3 $\pm$ 0.8 \\
OLMo-2-7B & KodCode & HumanEval & \textbf{34.5 $\pm$ 2.2} & 32.1 $\pm$ 2.2 & 32.8 $\pm$ 2.4 & 33.3 $\pm$ 1.8 & 33.4 $\pm$ 2.3 & 33.7 $\pm$ 2.1 \\
\bottomrule
\end{tabular}
}
}
\vspace{1em}
\caption{Scoring function design ablation results with $N=4$.}
\label{tab:ablation_n4}
\end{table}

\begin{table}[H]
\centering

\resizebox{\linewidth}{!}{%
{
\footnotesize
\setlength{\tabcolsep}{4pt}
\begin{tabular}{lll|ccccc|c}
\toprule
\textbf{Model} & \textbf{Train} & \textbf{Test} & \textbf{Final} & \textbf{Last-10} & \textbf{Special} & \textbf{Ensemble} & \textbf{Best Layer} & \textbf{\method} \\
\midrule
Qwen-30B-A3B & HumanEval & KodCode & 87.6 $\pm$ 0.6 & 86.8 $\pm$ 1.1 & \textbf{87.8 $\pm$ 0.6} & 87.4 $\pm$ 0.5 & 86.7 $\pm$ 0.9 & 87.7 $\pm$ 0.5 \\
Qwen-30B-A3B & KodCode & HumanEval & 90.4 $\pm$ 0.9 & 90.5 $\pm$ 0.9 & \textbf{91.0 $\pm$ 0.7} & 90.6 $\pm$ 0.9 & 90.9 $\pm$ 1.0 & 90.2 $\pm$ 0.8 \\
\midrule
Qwen-1.7B & HumanEval & KodCode & \textbf{62.9 $\pm$ 14.9} & 62.6 $\pm$ 14.4 & 58.1 $\pm$ 14.2 & 61.5 $\pm$ 15.9 & 62.0 $\pm$ 15.9 & 62.4 $\pm$ 14.8 \\
Qwen-1.7B & KodCode & HumanEval & 53.0 $\pm$ 9.4 & 52.4 $\pm$ 8.7 & 51.5 $\pm$ 6.6 & \textbf{53.9 $\pm$ 9.8} & 53.8 $\pm$ 9.6 & 52.6 $\pm$ 9.4 \\
\midrule
GPT-OSS-20B & HumanEval & KodCode & 86.5 $\pm$ 1.0 & 85.9 $\pm$ 0.5 & 85.7 $\pm$ 1.7 & 86.3 $\pm$ 2.2 & 85.5 $\pm$ 2.0 & \textbf{86.8 $\pm$ 0.8} \\
GPT-OSS-20B & KodCode & HumanEval & 92.3 $\pm$ 0.9 & 92.1 $\pm$ 0.7 & 91.9 $\pm$ 1.4 & 92.6 $\pm$ 0.9 & \textbf{92.7 $\pm$ 0.7} & 92.4 $\pm$ 0.9 \\
\midrule
OLMo-2-7B & HumanEval & KodCode & 32.2 $\pm$ 0.9 & 29.6 $\pm$ 1.1 & 29.2 $\pm$ 1.2 & \textbf{33.2 $\pm$ 1.3} & 32.2 $\pm$ 1.0 & 32.2 $\pm$ 0.9 \\
OLMo-2-7B & KodCode & HumanEval & \textbf{35.7 $\pm$ 1.6} & 33.4 $\pm$ 1.5 & 31.9 $\pm$ 2.0 & 35.3 $\pm$ 2.2 & 34.9 $\pm$ 1.7 & 35.2 $\pm$ 2.0 \\
\bottomrule
\end{tabular}
}
}
\vspace{1em}
\caption{Scoring function design ablation results with $N=8$.}
\label{tab:ablation_n8}
\end{table}

\begin{table}[H]
\centering

\resizebox{\linewidth}{!}{%
{
\footnotesize
\setlength{\tabcolsep}{4pt}
\begin{tabular}{lll|ccccc|c}
\toprule
\textbf{Model} & \textbf{Train} & \textbf{Test} & \textbf{Final} & \textbf{Last-10} & \textbf{Special} & \textbf{Ensemble} & \textbf{Best Layer} & \textbf{\method} \\
\midrule
Qwen-30B-A3B & HumanEval & KodCode & 87.8 $\pm$ 0.7 & 86.7 $\pm$ 1.3 & 87.8 $\pm$ 0.7 & 87.5 $\pm$ 0.7 & 86.8 $\pm$ 1.1 & \textbf{88.0 $\pm$ 0.6} \\
Qwen-30B-A3B & KodCode & HumanEval & 90.8 $\pm$ 0.7 & 91.1 $\pm$ 0.5 & 91.0 $\pm$ 0.8 & 90.7 $\pm$ 0.9 & \textbf{91.1 $\pm$ 0.8} & 90.6 $\pm$ 0.8 \\
\midrule
Qwen-1.7B & HumanEval & KodCode & \textbf{65.8 $\pm$ 13.0} & 64.7 $\pm$ 12.9 & 60.4 $\pm$ 13.6 & 63.9 $\pm$ 14.7 & 64.6 $\pm$ 14.6 & 65.2 $\pm$ 13.1 \\
Qwen-1.7B & KodCode & HumanEval & 57.6 $\pm$ 7.9 & 56.2 $\pm$ 7.7 & 55.5 $\pm$ 6.1 & \textbf{58.4 $\pm$ 8.5} & 57.6 $\pm$ 8.3 & 56.8 $\pm$ 8.6 \\
\midrule
GPT-OSS-20B & HumanEval & KodCode & 86.5 $\pm$ 1.2 & 85.8 $\pm$ 0.8 & 85.7 $\pm$ 1.7 & 86.3 $\pm$ 2.3 & 85.3 $\pm$ 2.6 & \textbf{86.5 $\pm$ 0.8} \\
GPT-OSS-20B & KodCode & HumanEval & 92.7 $\pm$ 0.8 & 92.0 $\pm$ 1.1 & 92.2 $\pm$ 0.9 & \textbf{93.1 $\pm$ 0.5} & 93.0 $\pm$ 0.8 & 93.0 $\pm$ 0.8 \\
\midrule
OLMo-2-7B & HumanEval & KodCode & 32.8 $\pm$ 1.0 & 30.4 $\pm$ 0.9 & 29.5 $\pm$ 1.4 & \textbf{33.5 $\pm$ 1.2} & 32.5 $\pm$ 1.1 & 32.8 $\pm$ 0.9 \\
OLMo-2-7B & KodCode & HumanEval & 35.2 $\pm$ 2.1 & 33.9 $\pm$ 1.8 & 30.9 $\pm$ 2.2 & 35.2 $\pm$ 1.9 & 35.1 $\pm$ 2.3 & \textbf{35.3 $\pm$ 1.2} \\
\bottomrule
\end{tabular}
}
}
\vspace{1em}
\caption{Scoring function design ablation results with $N=12$.}

\label{tab:ablation_n12}
\end{table}

\subsection{Additional Experiments}
\label{app:additional_experiments}

\subsubsection{LoRA - Domain Transfer Experiments}
\autoref{tab:transfer} compares the performance of the scoring model under in-domain and out-of-domain training settings, where LoRA is used to adapt the base model into a scoring model trained for trajectory evaluation. In the in-domain setting, the scoring model is trained and evaluated on the same domain (either mathematical reasoning or coding), whereas in the out-of-domain setting, the model is trained on one domain (e.g., math) and evaluated on the other (e.g., coding), and vice versa. We report accuracy across both settings to assess the extent to which the learned scoring function generalizes across domains. While transfer performance is often competitive, indicating that the model captures some domain-agnostic signals, in-domain training consistently yields stronger results.
\begin{table}[h]
\centering
\scriptsize
\setlength{\tabcolsep}{4pt}
\renewcommand{\arraystretch}{1.1}
\begin{tabular}{llcccc}
\toprule
Model & Test & Train (out-of-domain) & Train (in-domain) & Out-of-domain Acc. & In-domain Acc.  \\
\midrule

\multirow{8}{*}{GPT-OSS-20B}
 & HumanEval & AIME & KodCode & 91.0 $\pm$ 1.5 & 92.3 $\pm$ 0.4 \\
 & KodCode & AIME & HumanEval & 83.8 $\pm$ 0.9 & 85.2 $\pm$ 1.0 \\
 & AIME & HumanEval & MATH-500 & 94.4 $\pm$ 3.8 & 94.8 $\pm$ 2.4 \\
 & MATH-500 & HumanEval & AIME & 94.9 $\pm$ 0.5 & 94.8 $\pm$ 0.3 \\
 & AIME & KodCode & MATH-500 & 92.2 $\pm$ 3.8 & 94.8 $\pm$ 2.4 \\
 & MATH-500 & KodCode & AIME & 94.9 $\pm$ 0.9 & 94.8 $\pm$ 0.3 \\
 & HumanEval & MATH-500 & KodCode & 91.0 $\pm$ 1.2 & 92.3 $\pm$ 0.4 \\
 & KodCode & MATH-500 & HumanEval & 84.4 $\pm$ 1.3 & 85.2 $\pm$ 1.0 \\

\midrule
\multirow{4}{*}{OLMo-2-7B}
 & HumanEval & AIME & KodCode & 35.6 $\pm$ 2.7 & 37.7 $\pm$ 2.7 \\
 & KodCode & AIME & HumanEval & 28.3 $\pm$ 0.7 & 32.7 $\pm$ 0.8 \\
 & HumanEval & MATH-500 & KodCode & 35.2 $\pm$ 2.4 & 37.7 $\pm$ 2.7 \\
 & KodCode & MATH-500 & HumanEval & 30.9 $\pm$ 0.9 & 32.7 $\pm$ 0.8 \\

\midrule
\multirow{8}{*}{Qwen-1.7B}
 & HumanEval & AIME & KodCode & 49.6 $\pm$ 9.2 & 63.5 $\pm$ 9.1 \\
 & KodCode & AIME & HumanEval & 57.9 $\pm$ 19.5 & 68.3 $\pm$ 9.0 \\
 & AIME & HumanEval & MATH-500 & 43.0 $\pm$ 5.1 & 50.4 $\pm$ 5.9 \\
 & MATH-500 & HumanEval & AIME & 88.6 $\pm$ 0.5 & 88.9 $\pm$ 0.5 \\
 & AIME & KodCode & MATH-500 & 45.9 $\pm$ 3.6 & 50.4 $\pm$ 5.9 \\
 & MATH-500 & KodCode & AIME & 89.0 $\pm$ 0.5 & 88.9 $\pm$ 0.5 \\
 & HumanEval & MATH-500 & KodCode & 51.3 $\pm$ 7.8 & 63.5 $\pm$ 9.1 \\
 & KodCode & MATH-500 & HumanEval & 59.2 $\pm$ 18.9 & 68.3 $\pm$ 9.0 \\

\bottomrule
\end{tabular}
\vspace{1em}
\caption{Transfer vs.\ in-domain performance for finetuned full model using LoRA. We report accuracy when the scoring model is trained on an out-of-domain dataset versus an in-domain dataset. While transfer performance is often competitive, in-domain training consistently yields stronger results.}
\label{tab:transfer}
\end{table}

\subsubsection{Additional ReasonFlux Experiments}
\label{sec:reasonfluxexp}

We compare \method{} against process reward models (PRMs), which evaluate intermediate reasoning steps rather than only final answers and are commonly used to improve solution selection on challenging reasoning tasks. In particular, we compare against ReasonFlux, a recent PRM-based approach for modeling the quality of reasoning trajectories. Results on Qwen-14B across three mathematical reasoning benchmarks are reported in \autoref{tab:scatr_vs_reasonflux2}. \method{} consistently outperforms ReasonFlux-1.5B, despite being trained on only 1K OpenThoughts examples (500 math and 500 code), whereas ReasonFlux is trained on the full 114K-example dataset. These results underscore the strong data efficiency of \method{}, showing that a lightweight scorer trained on a small calibration set can match or exceed substantially more data-intensive PRM approaches.

\method{} also remains competitive with ReasonFlux-7B. On AIME24 and AIME25, its performance lies within one standard deviation of the larger model, with an absolute gap of only around 5\% on MATH-500. This is especially notable given the substantial disparity in scale: our scorer has roughly 2M parameters and is trained on just 1K examples, whereas ReasonFlux-7B has 7B parameters and is trained on 114K examples. Training and deploying a PRM at this scale is considerably more expensive, requiring substantially greater compute, memory, and supervision data than our lightweight scorer. While larger PRMs remain valuable when the goal is to maximize absolute performance, these results suggest a compelling cost--quality tradeoff in which much smaller scorers can achieve comparable performance on many problems at a fraction of the training and inference cost.

More broadly, this comparison suggests a practical hybrid design: a small, inexpensive scorer such as our MLP could handle routine or lower-ambiguity problems, while a larger PRM could be reserved for more challenging cases where additional reasoning capacity justifies the extra computational cost.

\begin{table}[h]
\centering
\small
\begin{tabular}{lccc}
\toprule
Dataset & \method~ & ReasonFlux-1.5B & ReasonFlux-7B \\
\midrule
AIME24   & 18.5$\pm$4.5 & 17.8$\pm$1.9 & 24.4$\pm$1.9 \\
AIME25     & 21.6$\pm$5.5 & 14.4$\pm$1.9 & 23.3$\pm$3.3 \\
MATH-500 & 80.4$\pm$0.7 & 79.2$\pm$0.4 & 85.7$\pm$0.5 \\
\bottomrule
\end{tabular}
\vspace{1em}
\caption{Comparison of \method~ with ReasonFlux models across mathematical reasoning benchmarks using Qwen2.5-14B model for data generation, following \cite{zou2025reasonfluxprmtrajectoryawareprmslong}. \method~ consistently outperforms the ReasonFlux-1.5B model despite being trained on only 1K OpenThoughts problems (randomly sampled with 500 code and 500 math examples), whereas ReasonFlux models are trained on the full 114K OpenThoughts dataset. ReasonFlux-7B dominates performance, but is more expensive to train and does not offer the same lightweight accuracy-efficiency tradeoff that~\method~does.}
\label{tab:scatr_vs_reasonflux2}
\end{table}

\subsubsection{Domain-Adaptive Inference}
Here, we discuss two methods to make \method~ domain-adaptive. In the first approach, we train a classifier on a calibration set formed from problems of both domains. This classifier can be applied to rollouts of both math and coding. In the second, we use the LLM to classify the problem as math or coding (limiting output to five tokens). Then, we route the problem's rollouts to the corresponding classifier.

We construct mixed calibration sets by combining two domains with a 50--50 task split, and train a single classifier on this pooled data using embeddings from the same layer for both domains. Results suggest that \method~ does not require a highly domain-robust calibration set to perform well, and that single-domain calibration is generally sufficient.

We also investigate a two-stage routing system that first classifies each problem by domain before selecting the appropriate classifier. Rather than training a single classifier on a mixed calibration set (as in the third ablation), we prepend an LLM-based router: given a problem statement, the same LLM used to generate responses is prompted to label the question as either \texttt{math} or \texttt{coding}, and the input is then forwarded to the corresponding single-domain \method~ classifier trained exclusively on that domain. We classified approximately 3{,}000 questions spanning all five evaluation datasets (\textsc{HumanEval}, \textsc{KodCode}, \textsc{BigCodeBench-Hard}, \textsc{MATH500}, and \textsc{AIME}) and found that the LLM router achieves $100\%$ classification accuracy after prompt refinement, introducing no routing errors into the pipeline. This allows us to isolate the question of whether domain-specialized classifiers, when correctly routed to, outperform both the mixed-calibration classifier and the single-domain baselines. The two-stage system, therefore, represents the best-case deployment scenario for \method~ in a setting where problems may be drawn from heterogeneous domains, and the results speak to whether the performance gap between single-domain and mixed-domain calibration can be recovered through routing rather than requiring separate inference pipelines per domain from the outset.

Results for the mixed calibration set approach and the two-stage routing 
approach are shown in \autoref{tab:code_mixed_main}.

\begin{table}[h]
\centering
\resizebox{\linewidth}{!}{%
{
\small
\setlength{\tabcolsep}{4pt}
\begin{tabular}{llll|cc}
\toprule
\textbf{Model} & \textbf{Train (mixed)} & \textbf{Train (two-stage)} & \textbf{Test} & \textbf{Mixed-domain} & \textbf{Two-stage routing}\\
\midrule
GPT-OSS-20B & MATH-500+HumanEval & Humaneval & KodCode & 86.2 $\pm$ 0.9 & 86.8 $\pm$ 0.7 \\
GPT-OSS-20B & MATH-500+KodCode & KodCode & HumanEval & 92.8 $\pm$ 0.7 & 93.2 $\pm$ 0.6 \\
\midrule
Qwen-30B-A3B & MATH-500+HumanEval & Humaneval & KodCode & 88.0 $\pm$ 0.7 & 88.3 $\pm$ 0.6 \\
Qwen-30B-A3B & MATH-500+KodCode & KodCode & HumanEval & 90.8 $\pm$ 0.7 & 90.4 $\pm$ 0.8 \\
\midrule
OLMo-2-7B & MATH-500+HumanEval & Humaneval & KodCode & 33.2 $\pm$ 1.3 & 33.1 $\pm$ 0.9  \\
OLMo-2-7B & MATH-500+KodCode & KodCode & HumanEval & 34.5 $\pm$ 2.1 & 35.0 $\pm$ 1.7  \\
\bottomrule
\end{tabular}
}
}
\vspace{1em}
\caption{We compare mixed-domain training to a two-stage routing approach. We compare Best-of-$N$ accuracy where $N=16$. We see that mixing domains does not sacrifice performance, and therefore, just a one classifier trained on math and coding data may be sufficient.}
\label{tab:code_mixed_main}
\end{table}

\subsubsection{\method~: In-distribution vs. out-of-distribution results}
We study whether using the same dataset to train and evaluate \method~ leads to improvement. We use a 50-50 train/test split, evaluating Best-of-$N$ accuracy on the held-out 50\%. Results are shown in \autoref{tab:same_domain_main}.

\begin{table}[H]
\centering

\resizebox{.6\linewidth}{!}{%
{
\small
\setlength{\tabcolsep}{4pt}
\begin{tabular}{ll|cc}
\toprule
\textbf{Model} & \textbf{Train Dataset} & \textbf{Same test dataset} & \textbf{Different test dataset} \\
\midrule
Qwen-30B-A3B & HumanEval & 88.5 $\pm$ 1.5 & 90.4 $\pm$ 0.8 \\
Qwen-30B-A3B & KodCode & 87.1 $\pm$ 1.3 & 88.3 $\pm$ 0.6 \\
Qwen-30B-A3B & MATH-500 & 89.1 $\pm$ 0.4 & 91.3 $\pm$ 0.5 \\
\midrule
GPT-OSS-20B & HumanEval & 92.2 $\pm$ 1.2 & 93.2 $\pm$ 0.6 \\
GPT-OSS-20B & KodCode & 87.4 $\pm$ 1.2 & 86.8 $\pm$ 0.7 \\
GPT-OSS-20B & MATH-500 & 96.1 $\pm$ 1.0 & 95.0 $\pm$ 0.5 \\
\midrule
Qwen-1.7B & HumanEval & 60.5 $\pm$ 9.9 & 59.3 $\pm$ 8.1 \\
Qwen-1.7B & KodCode & 67.8 $\pm$ 9.2 & 67.1 $\pm$ 12.2 \\
Qwen-1.7B & MATH-500 & 89.2 $\pm$ 1.1 & 88.3 $\pm$ 0.7 \\
\midrule
OLMo-2-7B & HumanEval & 29.8 $\pm$ 3.0 & 35.0 $\pm$ 1.7 \\
OLMo-2-7B & KodCode & 33.6 $\pm$ 1.6 & 33.1 $\pm$ 0.9 \\
OLMo-2-7B & MATH-500 & 34.6 $\pm$ 1.5 & 32.8 $\pm$ 1.2 \\
\bottomrule
\end{tabular}
}
}
\vspace{1em}
\caption{Results for training and evaluating \method~ on the same dataset, compared to having been trained on a different dataset (as in the main results). Best-of-$N$ results are shown for $N=16$. Results are mostly similar, differing more in cases where the sizes of the datasets are different. The different datasets are as follows: for HumanEval, we train on Kodcode and vice versa. For Math-500, we train on AIME.}
\label{tab:same_domain_main}
\end{table}

\end{document}